\documentclass{article} 
\usepackage{iclr2024_conference,times}

\usepackage{amsmath,amsfonts,bm}

\def\eqref#1{equation~\ref{#1}}

\def\1{\bm{1}}

\DeclareMathAlphabet{\mathsfit}{\encodingdefault}{\sfdefault}{m}{sl}
\SetMathAlphabet{\mathsfit}{bold}{\encodingdefault}{\sfdefault}{bx}{n}

\usepackage{hyperref}
\usepackage{url}

\usepackage{booktabs}       
\usepackage{amsfonts}       
\usepackage{nicefrac}       
\usepackage{microtype}      
\usepackage{xcolor}         

\usepackage{makecell}
\usepackage{algorithm}
\usepackage{algorithmic}
\usepackage{graphicx}
\usepackage{caption}
\usepackage{wrapfig}
\usepackage{subcaption}
\usepackage{amsmath}
\usepackage{amssymb}
\usepackage{mathtools}
\usepackage{amsthm}

\newcommand{\etc}{etc}
\usepackage[capitalize,noabbrev]{cleveref}
\newtheorem{theorem}{Theorem}
\newtheorem{corollary}{Corollary}
\newtheorem{assumption}{Assumption}
\newtheorem{proposition}{Proposition}
\newtheorem{hypothesis}{Hypothesis}
\newtheorem{axiom}{Candidate Axiom}

\usepackage{listings}
\usepackage{xcolor}

\newcommand\indep{\protect\mathpalette{\protect\independenT}{\perp}}
\def\independenT#1#2{\mathrel{\rlap{$#1#2$}\mkern2mu{#1#2}}}

\title{Indeterminate Probability Theory}

\iclrfinalcopy

\author{Tao Yang,
Chuang Liu,
Xiaofeng Ma,
Weijia Lu,
Ning Wu,
Bingyang Li, \\
\textbf{
Zhifei Yang,
Peng Liu,
Lin Sun,
Xiaodong Zhang,
Can Zhang }
\\
AI Lab, United Automotive Electronic Systems Co., Ltd. \\
Shanghai, China \\
\texttt{tao.yang9@uaes.com}
}

\begin{document}

\maketitle

\begin{abstract}
Complex continuous or mixed joint distributions (e.g., $P(Y \mid z_1,z_2,\dots,z_N)$) generally lack closed-form solutions, often necessitating approximations such as MCMC. 
This paper proposes \textbf{Indeterminate Probability Theory} (IPT), which makes the following contributions:
(1) An \textbf{observer-centered framework} in which experimental outcomes are represented as distributions combining ground truth with observation error;  
(2) The introduction of three independence candidate \textbf{axioms} that enable a two-phase probabilistic inference framework; 
(3) The derivation of \textbf{closed-form solutions} for arbitrary complex joint distributions under this framework.
Both the \textit{Indeterminate Probability Neural Network (IPNN)} model and the non-neural \textit{multivariate time series forecasting} application demonstrate IPT's effectiveness in modeling high-dimensional distributions, with successful validation up to 1000 dimensions.
Importantly, IPT is consistent with classical probability theory and subsumes the frequentist equation in the limit of vanishing observation error.
\footnote{Source code: \url{https://github.com/Starfruit007/ipnn}}

\end{abstract}

\section{Introduction}\label{sec:introduction}
Classical probability theory, particularly its frequentist interpretation, relies on observing and counting outcomes across repeated trials. Under this framework, each event/sample is assumed to be clearly defined and unambiguously observed — a prerequisite for computing stable frequency estimates that converge to well-defined probabilities.

However, this assumption often fails in real-world scenarios. Observations are often ambiguous, observer-dependent, or constrained by measurement limitations. For instance, in a coin toss experiment, while we may expect to observe either heads or tails, imperfect visibility or limited resolution can lead to uncertainty about the actual outcome. In such cases, the very notion of a discrete, uniquely determined sample point becomes questionable.
More generally, observations may not only be uncertain but also exhibit continuous variability. Again, in a coin toss experiment, an observer may interpret the outcome as e.g. a Gaussian distribution based on their special concerns or measurement context. This suggests that a more general theory is needed — one that can accommodate both discrete and continuous forms of observation uncertainty within a unified framework.

To address this need, we propose \textbf{Indeterminate Probability Theory} (IPT), a new framework that extends classical probability by explicitly modeling the observer's role and the uncertainty inherent in the observation process. Unlike traditional models that treat observations as direct proxies for truth, IPT begins from the premise that all knowledge arises from observation outputs, which reflect both the underlying system and the conditions under which it is observed.

This perspective leads to a structured two-phase approach:
\begin{itemize}
   \item \textbf{Observation Phase:} Rigorously defining conditional relationships among observable outputs (discrete, continuous, or mixed) via candidate Axiom~\ref{axm:A_X_independence} and Axiom~\ref{axm:Y_A_independence}.
   \item \textbf{Inference Phase:} Performing probabilistic inference with imperfect observable outputs based on Axiom~\ref{axm:Y_X_independence}.
\end{itemize}

We demonstrate the utility of IPT through two practical applications:

\begin{itemize}
   \item\textbf{IPNN (Indeterminate Probability Neural Network):} A discrete/continuous neural architecture achieving tractable inference in high-dimensional latent spaces (up to 1000 dimensions); \cite{cipnn} 
   \item \textbf{Non-neural multivariate time series forecasting:} IPT-based method outperforms LSTM and Transformer baselines by modeling observer-induced uncertainty.  \cite{seqip}
\end{itemize}

Importantly, IPT is not at odds with classical probability theory. Instead, it subsumes frequentist probability as a special case when observational error vanishes (Theorem~\ref{thm:classical_subsumption}), ensuring compatibility with existing methodologies. By bridging the gap between theoretical rigor and practical robustness, IPT offers a unified framework for probabilistic reasoning in uncertain environments.


\section{A Toy Example}\label{sec:example}

To illustrate the concept of indeterminate probability theory (IPT), we present a coin toss experiment with three distinct observers.
This scenario demonstrates how IPT resolves questions intractable to classical probability theory when observation uncertainty exists. 
Experimental parameters are detailed in Table~\ref{tab:cipnn_example_background}.

\begin{table}[htbp]
    \caption{Coin Toss Experiment}
    \label{tab:cipnn_example_background}
    \begin{center}
    \begin{tabular}{cccccc}
    \hline
    Random Experiment ID $X$ & $x_{1}$ & $x_{2}$ &  $x_{3}$ &  $x_{4}$ &  $x_{5}$ \\
    &  $x_{6}$ &  $x_{7}$ &  $x_{8}$ &  $x_{9}$ &  $x_{10}$ \\ 
    \\[-1.5ex]
    Truth & $hd$ & $hd$ & $hd$ & $hd$ & $hd$ \\
    & $tl$ & $tl$ & $tl$ & $tl$ & $tl$ \\
    \\[-1.5ex]
    Record of Observer$_{1}$ $Y$ & $hd$ & $hd$ & $hd$ & $hd$ & $hd$ \\
    & $tl$ & $tl$ & $tl$ & $tl$ & $tl$ \\
    \\[-1.5ex]
    Equivalent Record $Y$ & 1, 0 & 1, 0 & 1, 0 & 1, 0 & 1, 0 \\
    & 0, 1 & 0, 1 & 0, 1 & 0, 1 & 0, 1 \\
    \\[-1.5ex]
    Record of Observer$_{2}$  $A$ & 0.8, 0.2 & 0.7, 0.3 & 0.9, 0.1 & 0.6, 0.4 & 0.8, 0.2 \\
    & 0.1, 0.9 & 0.2, 0.8 & 0.3, 0.7 & 0.1, 0.9 & 0.2, 0.8 \\
    \\[-1.5ex]
    Record of Observer$_{3}$ $z$ & $\mathcal{N}(3,1)$ & $\mathcal{N}(3,1)$ & $\mathcal{N}(3,1)$ & $\mathcal{N}(3,1)$ & $\mathcal{N}(3,1)$ \\
    & $\mathcal{N}(-3,1)$ & $\mathcal{N}(-3,1)$ & $\mathcal{N}(-3,1)$ & $\mathcal{N}(-3,1)$ & $\mathcal{N}(-3,1)$ \\  
    \hline
    \end{tabular}
    \end{center}
    Where $hd$ is for head, $tl$ is for tail. And conditioning on $x_{k}$ is the indeterminate probability, e.g.  $P(Y=hd|X=x_{3}) = 1$, $P(A=tl|X=x_{6}) = 0.9$ and $P(z|X=x_{8}) = \mathcal{N}(z;-3,1)$.
    \end{table}

\textbf{Observer\textsubscript{1}} records outcomes perfectly. The probability of heads is:
\begin{equation}
P(Y=hd)=\frac{\text{number of }(Y=hd)\text{ occurs}}{\text{number of random experiments}} = \frac{5}{10}
\end{equation}

By defining the experiment ID as a random variable $X$, we can also represent Observer$_{1}$'s record with equivalent form of $P(Y=hd|X=x_{k})$, lead to
\begin{equation}
        P(Y=hd)=\sum_{k=1}^{10}P(Y=hd|X=x_{k})\cdot P(X=x_{k}) = \frac{5}{10}  
\end{equation}

Note that random variable $X$ is special, only condition on $X$ has a special meaning for the observation of each coin toss.

\textbf{Observer\textsubscript{2}} outputs probability distributions. The head probability is:

\begin{equation}
    P(A=hd)=\sum_{k=1}^{10}P(A=hd|X=x_{k})\cdot P(X=x_{k}) = \frac{4.7}{10}
\end{equation}

This combines \textbf{ground truth} and \textbf{observation error}.

\textbf{Observer\textsubscript{3}} outputs Gaussian distributions $\mathcal{N}(z;\mu,1)$ with unknown mapping. 
The distribution is:

\begin{equation}
P(z)=\sum_{k=1}^{10}P(z|X=x_{k})\cdot P(X=x_{k})  
= \frac{5\cdot\mathcal{N}(z;3,1)+5\cdot\mathcal{N}(z;-3,1)}{10}
\end{equation}

The bimodal $P(z)$ raises a key question: How do we mathematically associate each mode with physical outcomes? 
Classical probability cannot resolve this in closed-form.

Using IPT's conditional independence Axiom~\ref{axm:Y_A_independence} (given X, z and Y is conditional independent in observation phase):

\begin{equation}
    \begin{aligned}
    P(Y=hd|z) = \frac{\sum_{k=1}^{10}P(Y=hd|X=x_{k})\cdot P(z|X=x_{k})}{\sum_{k=1}^{10}P(z|X=x_{k})} = \frac{\mathcal{N}(z;3,1)}{\mathcal{N}(z;3,1)+\mathcal{N}(z;-3,1)}
    \end{aligned}
\end{equation}

For a new toss $X_{11}$ with $P(z|X=x_{11})=\mathcal{N}(z;3,1)$, applying inference-phase independence Axiom~\ref{axm:Y_X_independence} (given z, X and Y is conditional independent in inference phase), along with Monte Carlo method:

\begin{equation}
   \label{eq:example_gauss_coin_toss}
    \begin{aligned}
    P^z(Y=hd|X=x_{11}) &= \int_{z} \left( P(Y=hd|z, X=x_{11})\cdot P(z|X=x_{11}) \right)\\
    &= \int_{z} \left( P(Y=hd|z)\cdot P(z|X=x_{11}) \right)\\
    &= \mathbb{E}_{z \sim P(z|X=x_{11})}\left [ P(Y=hd|z)  \right ] \approx \frac{1}{C}\sum_{c=1}^{C}P(Y=hd|z_{c}) \\
    &= \frac{1}{C}\sum_{c=1}^{C}\frac{\mathcal{N}(z_{c};3,1)}{\mathcal{N}(z_{c};3,1)+\mathcal{N}(z_{c};-3,1)} \approx 1,  z_{c} \sim \mathcal{N}(z;3,1)
    \end{aligned}
\end{equation}

Where superscript $P^z(Y=hd|X=x_{11})$ indicates that the inference is based on the latent variables $z$,
and $P(Y=hd|X=x_{11})$ indicates that the record of the observer$_{3}$.
$C$ represents the number of Monte Carlo samples. 
This identifies $\mathcal{N}(z;3,1)$ with heads.

\textbf{Extensions}: When Observer\textsubscript{3} is a neural network outputting multivariate Gaussians, this yields the CIPNN model~\cite{cipnn}. Directly modeling time series as Gaussians (without neural networks) gives the forecasting method~\cite{seqip}.

\section{Indeterminate Probability Theory}\label{sec:ip_theory}

Let $A^{1},A^{2},...,A^{N}$ and $Y$ denote distinct discrete, continuous or mixed random variables. For simplicity,
we present the theory using discrete random variables, though the framework applies equally to continuous or mixed cases.

Current methods lack general analytical solutions for complex conditional distributions $P  (Y=y_{l}\mid A^{1}=a_{i_{1}}^{1},\dots, A^{N}=a_{i_{N}}^{N}) $
(compactly written as $P \left (y_{l} |a_{i_{1}}^1,a_{i_{2}}^2,\dots,a_{i_{N}}^N \right )$\footnote{Compact notation used throughout; multivariate $Y$ is permitted}). 
Indeterminate probability theory addresses this gap.

\subsection{Definition of Indeterminate Probability}

Define a special random variable $X$ to represent the i.i.d. random experiments, where $X=x_{k}$ corresponds to the $k^{th}$ experiment:

\begin{equation}
   \label{eq:P_X}    
   P\left ( x_{k}   \right ) = \frac{1}{n}, k = 1,2,\dots,n. 
   \end{equation}

As discussed in Section~\ref{sec:introduction}, observations (by machines, models, or humans) yield probability distributions for each experiment. 
Indeterminate probability represents the observed outcome of the $k^{th}$ experiment as

\begin{equation}
   \label{eq:P_a_X}    
   \text{Indeterminate Probability}:= P\left (a_{i_{j} }^{j}\mid x_{k}   \right ) \in [0,1]
   \end{equation}

Conditioning on $X$ has a distinct interpretation:   
$P(A|X=x_k)$ signifies the likelihood of $A$ occurring in the $k^{th}$ experiment.
This differs fundamentally from conditioning on other variables.

In classical probability, event states are binary: $P(A^j = a_{i_j}^j \mid X = x_k) \in \{0,1\}$.
For example (Section~\ref{sec:example}), $P(Y = hd \mid X = x_3) = 1$.
This distinction renders frequency-based equations inapplicable.

For multivariate variables $\mathbb{A} = \left (A^{1},A^{2},\dots,A^{N} \right )$,
observations from different observers are independent.
Empirical evidence suggests that this independence also holds for the same observer  
considering $Y$ and $A^{1},A^{2},...,A^{N}$ from different perspectives. We have Axiom~\ref{axm:A_X_independence}:

\begin{axiom}
   \label{axm:A_X_independence}
   $A^{1} \indep A^{2}\indep,\dots, A^{N} \mid X : $
   Given $X$, $A^{1},A^{2},\dots,A^{N}$ are conditionally mutually independent.
   \end{axiom}

The joint indeterminate probability is

\begin{equation}
   \label{eq:P_A_X}  
   P\left (a_{i_{1}}^{1}, a_{i_{2}}^{2},\dots, a_{i_{N}}^{N}\mid x_{k}   \right ) 
   =   \prod_{j=1}^{N}  P\left ( a_{i_{j} }^{j}\mid x_{k}   \right ) \in [0,1]
   \end{equation}

Where it can be easily proved,
\begin{equation}
   \label{eq:Sum_P_A_X}    
   \sum_{\mathbb{A}} \prod_{j=1}^{N}  P\left ( a_{i_{j} }^{j}\mid x_{k}   \right ) = 1, k=1,2,\dots ,n.
   \end{equation}

In classical probability, the joint indeterminate probability ${\textstyle \prod_{j=1}^{N}}P\left ( a_{i_{j} }^{j}\mid x_{k}   \right ) \in \{0,1\}$.

\subsection{Observation Phase}

The conditional probability is:

\begin{equation}
   \label{eq:P_Y_A_1}
   P\left (y_{l} \mid a_{i_{1}}^{1},a_{i_{2}}^{2},\dots,a_{i_{N}}^{N} \right ) 
   =  \frac{P\left (y_{l},  a_{i_{1}}^{1},a_{i_{2}}^{2},\dots,a_{i_{N}}^{N} \right )}
   {P\left ( a_{i_{1}}^{1},a_{i_{2}}^{2},\dots,a_{i_{N}}^{N} \right )} 
   \end{equation}

Using the total probability theorem over $X$ with Equation~\ref{eq:P_X} and Equation~\ref{eq:P_A_X}:

\begin{equation}
   \label{eq:P_A}
   \begin{aligned}
   P\left (a_{i_{1}}^{1},a_{i_{2}}^{2},\dots,a_{i_{N}}^{N} \right ) 
   &=   {\textstyle \sum_{k=1}^{n}}
   \left ( P\left (a_{i_{1}}^{1},a_{i_{2}}^{2},\dots,a_{i_{N}}^{N} \mid x_{k}\right )\cdot P(x_{k})  \right ) \\
   &=  {\textstyle \sum_{k=1}^{n}}
   \left ( {\textstyle \prod_{j=1}^{N}}P\left ( a_{i_{j} }^{j}\mid x_{k}   \right )  \cdot P(x_{k})  \right ) \\
   &=\frac{   {\textstyle \sum_{k=1}^{n}} \left ( {\textstyle \prod_{j=1}^{N}}P\left ( a_{i_{j} }^{j}\mid x_{k}   \right ) \right ) }{n} 
   \end{aligned}
   \end{equation}

Since $Y$ and $A^j$ derive from different observational perspectives (or same observer with different perspectives):

\begin{axiom}
   \label{axm:Y_A_independence}
   $Y \indep A^{j} \mid X : $
   Given $X$, $A^{j}$ and $Y$ are conditionally mutually independent in the observation phase, $j=1,2,\dots,N$.
   \end{axiom}

Thus:

\begin{equation}
   \label{eq:P_YA}
   \begin{aligned}
   P\left (y_{l}, a_{i_{1}}^{1},a_{i_{2}}^{2},\dots,a_{i_{N}}^{N} \right )
   &= {\textstyle \sum_{k=1}^{n}}
   \left ( P\left (y_{l}, a_{i_{1}}^{1},a_{i_{2}}^{2},\dots,a_{i_{N}}^{N} \mid x_{k}\right )\cdot P(x_{k})  \right ) \\
   &=  {\textstyle \sum_{k=1}^{n}}
   \left ( P\left (y_{l}\mid x_{k}\right )\cdot {\textstyle \prod_{j=1}^{N}}P\left ( a_{i_{j} }^{j}\mid x_{k}   \right )  \cdot P(x_{k})  \right ) \\
   &=\frac{{\textstyle \sum_{k=1}^{n}}
   \left ( P\left (y_{l}\mid x_{k}\right )\cdot {\textstyle \prod_{j=1}^{N}}P\left ( a_{i_{j} }^{j}\mid x_{k}   \right ) \right )}{n} 
   \end{aligned} 
\end{equation}

Substitute Equation~\ref{eq:P_A} and Equation~\ref{eq:P_YA} into Equation~\ref{eq:P_Y_A_1}:

\begin{equation}
   \label{eq:P_Y_A_2}
   P\left (y_{l}| a_{i_{1}}^{1},a_{i_{2}}^{2},\dots,a_{i_{N}}^{N} \right )
   = \frac{{\textstyle \sum_{k=1}^{n}} \left ( P\left (y_{l}\mid x_{k}\right )\cdot {\textstyle \prod_{j=1}^{N}}P\left ( a_{i_{j} }^{j}\mid x_{k}   \right ) \right )} 
   {  {\textstyle \sum_{k=1}^{n}} \left ( {\textstyle \prod_{j=1}^{N}}P\left ( a_{i_{j} }^{j}\mid x_{k}   \right ) \right ) }
   \end{equation}

Where it can be proved, 

\begin{equation}
   \label{eq:sum_P_A}
   {\textstyle \sum_{l=1}^{m}}P\left (y_{l} \mid a_{i_{1}}^{1},a_{i_{2}}^{2},\dots,a_{i_{N}}^{N} \right ) = 1  
   \end{equation}

Equation~\ref{eq:P_Y_A_2} provides an analytical solution for arbitrary conditional probabilities. 
When $P(a_{i_{j}}^{j}\mid x_{k}) \in\{0,1\}$ and $P(y_{l}\mid x_{k}) \in\{0,1\}$, it reduces to the classical frequency-based probability.

\subsection{Inference Phase}

\begin{figure}[htbp]
   \centering
         \begin{subfigure}[b]{0.3\linewidth}
               \centering
               \includegraphics[width=1\linewidth]{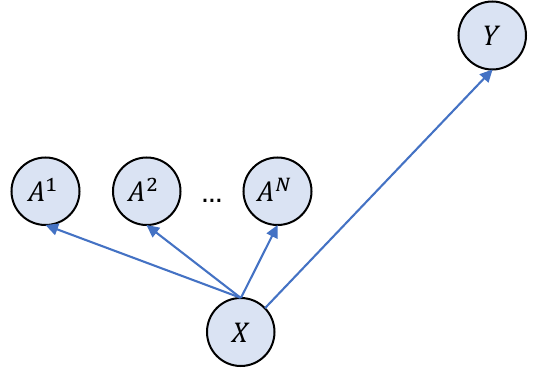}
               \caption{observation phase}
               \label{fig:observation}
         \end{subfigure}
         \hfil
         \begin{subfigure}[b]{0.3\linewidth}
               \centering
               \includegraphics[width=1\linewidth]{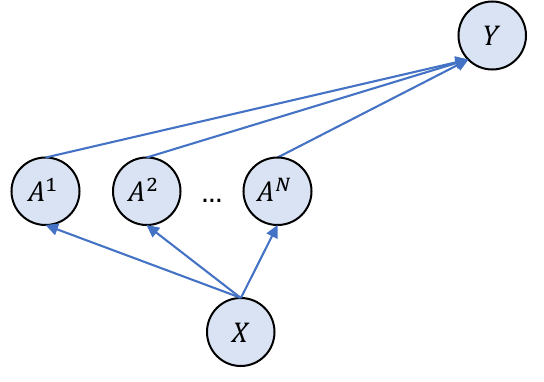}
               \caption{inference phase}
               \label{fig:inference}
         \end{subfigure}
         \caption{Independence illustration with Bayesian network.}
         \label{fig:probability_process}
   \end{figure}

Given $\mathbb{A}$ and using Equation~\ref{eq:P_Y_A_2} (based on passed experience), we can infer $Y=y_{l}$.
This inferred $y_{l}$ does not refer to any specific sample $x_{k}$, 
including new input sample $x_{n+1}$. We establish the following axiom:

\begin{axiom}
   \label{axm:Y_X_independence}
   $X \indep Y \mid \left (A^{1},A^{2},\dots,A^{N} \right ) : $
   Given $ \left (A^{1},A^{2},\dots,A^{N} \right )$, $X$ and $Y$ are conditionally mutually independent in the inference phase.
   \end{axiom}    

This phase distinction is necessary because $Y$ is unobserved for new $x_{n+1}$,
and avoids conflict between Axioms~\ref{axm:Y_A_independence} and~\ref{axm:Y_X_independence}.

For the next experiment $X=x_{n+1}$, by applying the total probability theorem 
over the joint sample space $\left (a_{i_{1}}^{1},a_{i_{2}}^{2},\dots,a_{i_{N}}^{N} \right ) \in \mathbb{A}$, 
and considering Axiom~\ref{axm:Y_X_independence}, Equation~\ref{eq:P_A_X} and Equation~\ref{eq:P_Y_A_2}, we derive the inference probability as

\begin{equation}
   \label{eq:P_Y_X_via_A}
   \begin{aligned}
   P^{\mathbb{A} } \left ( y_{l} \mid x_{n+1} \right ) &=\sum_{\mathbb{A}} 
   \left ( P\left ( y_{l},a_{i_{1}}^{1}, a_{i_{2}}^{2}  ,  \dots, a_{i_{N}}^{N} \mid x_{n+1} \right )  \right ) \\ 
   &=\sum_{\mathbb{A}} 
   \left ( P\left ( y_{l}\mid a_{i_{1}}^{1}, a_{i_{2}}^{2} ,  \dots, a_{i_{N}}^{N}\right ) \cdot P\left (a_{i_{1}}^{1}, a_{i_{2}}^{2} , \dots, a_{i_{N}}^{N} \mid x_{n+1} \right )  \right ) \quad \text{(Axiom~\ref{axm:Y_X_independence})}\\
   &=\sum_{\mathbb{A}} 
   \left (\frac{{\textstyle \sum_{k=1}^{n}} \left ( P\left (y_{l}\mid x_{k}\right )\cdot {\textstyle \prod_{j=1}^{N}}P\left ( a_{i_{j} }^{j}\mid x_{k}   \right ) \right )} 
   {  {\textstyle \sum_{k=1}^{n}} \left ( {\textstyle \prod_{j=1}^{N}}P\left ( a_{i_{j} }^{j}\mid x_{k}   \right ) \right ) } 
   \cdot\prod_{j=1}^{N}  P\left ( a_{i_{j} }^{j}\mid x_{n+1}   \right )  \right )
   \end{aligned}
\end{equation}

Where the superscript $\mathbb{A}$ denotes inference via latent variables $\mathbb{A}$.
$P^{\mathbb{A} } \left ( y_{l} \mid x_{n+1} \right )$ and $P\left ( y_{l} \mid x_{k} \right )$ are mathematically equivalent, representing inferred and observed indeterminate probabilities, respectively.

The discrete decision rule is:
\begin{equation}
   \label{eq:argmax}
   \hat{y} := { \underset{l\in \left \{ 1,2,\dots,m \right \}} {\arg\max} \,  P^{\mathbb{A} } \left ( y_{l} \mid x_{n+1} \right ) }  
   \end{equation}

\subsection{Phase Distinction}
   The framework operates in two distinct phases:
   
   \textbf{Observation Phase} builds probabilistic relationships exclusively from historical data $\mathcal{D} = \{(x_k, P(Y|x_k), P(\mathbb{A}|x_k))\}_{k=1}^n$. Under Axioms \ref{axm:A_X_independence} and \ref{axm:Y_A_independence}, it computes the core conditional distribution $P(Y|\mathbb{A})$ through Eq. \ref{eq:P_Y_A_2}. This phase requires complete distributional records $(P(Y|X), P(\mathbb{A}|X))$ for all $x_k \in \mathcal{D}$.
   
   \textbf{Inference Phase} utilizes $P(Y|\mathbb{A})$ for prediction on any i.i.d $x_t$ (including $x_t \notin \mathcal{D}$ or $x_t \in \mathcal{D}$). Given observer output $P(\mathbb{A}|X=x_t)$, it computes predictions via Eq. \ref{eq:P_Y_X_via_A} under Axiom \ref{axm:Y_X_independence}. Critically, this phase never modifies $P(Y|\mathbb{A})$ from the observation phase and
   treats $x_{\text{t}}$ as statistically independent of historical $Y$ given $\mathbb{A}$.

\subsection{Complexity Reduction}\label{sec:Complexity}

Equation~\ref{eq:P_Y_X_via_A} can be reformulated as an expectation.
Monte Carlo approximation reduces complexity from $O(m\prod_{j=1}^{N}M_{j})$ to $O(m n N C)$.

\begin{equation}
   \label{eq:P_Y_X_via_A_expectation}
   \begin{aligned}
   P^{\mathbb{A} } \left ( y_{l} \mid x_{n+1} \right )
   &=\mathbb{E}_{a_{i_{j} }^{j}\sim  P\left ( a_{i_{j} }^{j}\mid x_{n+1}   \right )}
   \left [\frac{{\textstyle \sum_{k=1}^{n}} \left ( P\left (y_{l}\mid x_{k}\right )\cdot {\textstyle \prod_{j=1}^{N}}P\left ( a_{i_{j} }^{j}\mid x_{k}   \right ) \right )} 
   {  {\textstyle \sum_{k=1}^{n}} \left ( {\textstyle \prod_{j=1}^{N}}P\left ( a_{i_{j} }^{j}\mid x_{k}   \right ) \right ) }  \right ] \\
   &\approx \frac{1}{C}\sum_{c=1}^{C} 
   \left (\frac{{\textstyle \sum_{k=1}^{n}} \left ( P\left (y_{l}\mid x_{k}\right )\cdot {\textstyle \prod_{j=1}^{N}}P\left ( a_{i_{j},c }^{j}\mid x_{k}   \right ) \right )} 
   {  {\textstyle \sum_{k=1}^{n}} \left ( {\textstyle \prod_{j=1}^{N}}P\left ( a_{i_{j},c }^{j}\mid x_{k}   \right ) \right ) }  \right ),
   \end{aligned}
\end{equation}

where $a_{i_{j},c}^{j}\sim  P\left ( a_{i_{j} }^{j}\mid x_{n+1}   \right )$ and $C$ represents the number of Monte Carlo samples.

Unlike Markov Chain Monte Carlo methods~\cite{monte_carlo}, which requires a large number of samples from a complex and high-dimensional space,
CIPNN achieves accurate results with $C=2$ even in 1000D latent spaces (see~\cite{cipnn}).

\subsection{Summary}\label{sec:summary}

\begin{theorem}[Frequency-based Probability Subsumption]\label{thm:classical_subsumption}
   When observation errors vanish such that all indeterminate probabilities become deterministic, i.e., 
   $P(a_{i_j}^j \mid x_k) \in \{0,1\}$ and $P(y_l \mid x_k) \in \{0,1\}$ 
   for all $j=1,\dots,N$, $l=1,\dots,m$, and $k=1,\dots,n, n+1$, 
   the inference probability $P^{\mathbb{A}}(Y=y_l \mid X=x_{n+1})$ in Equation~\ref{eq:P_Y_X_via_A} reduces to the classical frequency-based conditional probability.
   \end{theorem}
   
\begin{proof}
   Under deterministic observations:
   \begin{enumerate}
         \item The product $\prod_{j=1}^N P(a_{i_j}^j \mid x_k) \in \{0,1\}$ acts as an indicator function $\mathbb{I}_{\mathbb{A}}(x_k)$ for event $\mathbb{A}$ occurring in experiment $x_k$
         \item $P(y_l \mid x_k) \in \{0,1\}$ acts as indicator $\mathbb{I}_{y_l}(x_k)$ for outcome $y_l$ in $x_k$
         \item The observation-phase term simplifies to frequency counts:
         \[
         \frac{\sum_{k=1}^n \mathbb{I}_{y_l}(x_k) \cdot \mathbb{I}_{\mathbb{A}}(x_k)}{\sum_{k=1}^n \mathbb{I}_{\mathbb{A}}(x_k)} 
         = P_{\text{classical}}(y_l \mid \mathbb{A})
         \]
         \item For $x_{n+1}$, $\prod_{j=1}^N P(a_{i_j}^j \mid x_{n+1}) \in \{0,1\}$ selects the actual event $\mathbb{A}^*$ (1 when $\mathbb{A} = \mathbb{A}^*$, 0 otherwise)
         \item The inference sum collapses to the classical prediction:
         \[
         \sum_{\mathbb{A}} \left( P_{\text{classical}}(y_l \mid \mathbb{A}) \cdot \mathbb{I}_{\mathbb{A}}(x_{n+1}) \right) = P_{\text{classical}}(y_l \mid \mathbb{A}^*)
         \qedhere
         \]
   \end{enumerate}
   \end{proof}

Our core contribution is the tractable probability formulation:

\begin{align}   
&\boldsymbol{ P^{\mathbb{A}}\left ( Y=y_{l}\mid X=x_{n+1}\right )} \nonumber \\
\label{eq:eq1}
&=\sum_{\mathbb{A}}P\left ( y_{l},\mathbb{A}\mid x_{n+1}   \right ) \quad \text{(marginalization)}\\
\label{eq:eq2}
&=\sum_{\mathbb{A}}\left (P\left ( y_{l}\mid \mathbb{A} \right )
\cdot P(\mathbb{A}\mid x_{n+1}) \right ) \quad \text{(Axiom~\ref{axm:Y_X_independence})} \\
\label{eq:eq3}
&=\sum_{\mathbb{A}}\left (\frac{ {\textstyle \sum_{k=1}^{n}\left ( 
   P(y_{l}\mid x_{k})\cdot  P(\mathbb{A}\mid x_{k}) \right ) } } {{\textstyle \sum_{k=1}^{n}P(\mathbb{A}\mid x_{k})  } }
\cdot P(\mathbb{A}\mid x_{n+1}) \right ) \quad \text{(Axiom~\ref{axm:Y_A_independence})}\\
\label{eq:eq4}
&=\underset{\text{Inference phase}}{\underbrace{\sum\limits_{\mathbb{A}}\left (  
\underset{\text{Observation phase}}{\underbrace{\frac{ \sum\limits_{k=1}^{n}
   \left ( \boldsymbol{ P\left ( y_{l}\mid x_{k}   \right )}\cdot
\prod\limits_{j=1}^{N}P\left ( a_{i_{j} }^{j}\mid X=x_{k}   \right )   \right )}
{\sum\limits_{k=1}^{n}
   \left ( \prod\limits_{j=1}^{N}P\left ( a_{i_{j} }^{j}\mid x_{k}   \right )   \right )} }}
\cdot  \prod\limits_{j=1}^{N}P\left ( a_{i_{j} }^{j}\mid x_{n+1}   \right ) \right )}}   \quad \text{(Axiom~\ref{axm:A_X_independence})}
\end{align}

This formulation remains valid for continuous or mixed latent variables $\mathbf{z}$. In such cases, 
the summation $\sum_{\mathbb{A}}$ must be replaced by the appropriate integration:
\begin{itemize}
    \item For continuous $\mathbf{z}$: $\sum_{\mathbb{A}} \rightarrow \int_{\mathbf{z}} d\mathbf{z}$
    \item For mixed discrete-continuous $\mathbf{z}$: $\sum_{\mathbb{A}} \rightarrow \sum_{\mathbf{z}_{\text{disc}}} \int_{\mathbf{z}_{\text{cont}}} d\mathbf{z}_{\text{cont}}$
\end{itemize}

The three axioms of conditional independence are foundational but not formally provable.
Validation relies on empirical evidence, and we encourage counterexamples. 
Should even a toy dataset contradict these axioms, the validity of the proposed theory would be falsified.

Finally, Equation~\ref{eq:eq4} subsumes frequency-based probability as a special case when observation error vanishes
, as discussed in Theorem~\ref{thm:classical_subsumption}.
See Appendix~\ref{app:intuitive_explain} for intuition.

\section{Applications}\label{sec:applications}

\subsection{IPNN}\label{sec:ipnn}

For neural network tasks, $X=x_{k}$ is for the  $k^{th}$ input sample, $P(y_{l}|x_{k})=y_{l}(k)\in[0,1]$ is for the soft/hard label of train sample $x_{k}$, 
$P^{\mathbb{A}} \left ( y_{l} \mid x_{t} \right )$ is for the predicted label of test sample $x_{t}$.

\begin{figure}
  \centerline{\includegraphics[width=0.5\linewidth]{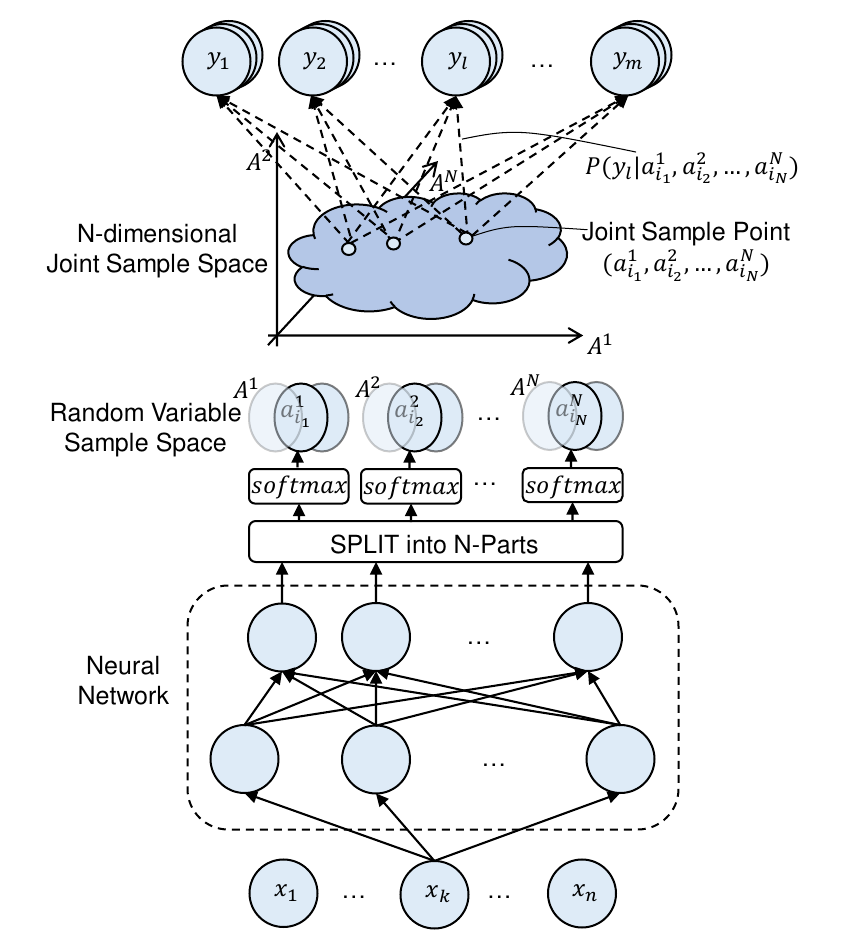}}
  \caption{IPNN model architecture. $P \left (y_{l} |a_{i_{1}}^1,a_{i_{2}}^2,\dots,a_{i_{N}}^N \right )$ is statistically calculated, not model weights.}
  \label{fig:ModelArchitecture}
  \end{figure}

Figure~\ref{fig:ModelArchitecture} shows IPNN model architecture, the output neurons of a general neural network 
(FFN, CNN, Resnet~\cite{resnet}, Transformer~\cite{Transformer}, Pretrained-Models~\cite{bert}, etc.) 
is split into N unequal/equal parts, the split shape is marked as Equation~\ref{eq:split_shape}, 
hence, the number of output neurons is the summation of the split shape ${\textstyle \sum_{j=1}^{N}M_{j}}$.
Next, each split part is passed to `softmax', so the output neurons can be defined as discrete random variable 
$A^{j} \in \left \{ a_{1}^{j} , a_{2}^{j} , \dots, a_{M_{j}}^{j} \right \},j=1,2,\dots,N$, and each neuron in $A^{j}$ is regarded as an event. 
After that, all the random variables together form the N-dimensional joint sample space, marked as $\mathbb{A} = (A^{1},A^{2},\dots,A^{N})$, and all the joint sample points are fully connected 
with all labels $Y \in \{y_{1},y_{2},\dots,y_{m} \}$ via conditional probability $P \left (y_{l} |a_{i_{1}}^1,a_{i_{2}}^2,\dots,a_{i_{N}}^N \right )$.

\begin{equation}
   \label{eq:split_shape}    
   \text{Split shape} := \{ M_{1},M_{2},\dots,M_{N} \} 
   \end{equation}

Given an input sample $x_{k}$, let $\alpha _{i_{j}}^{j}(k)$ be the model outputted value after `softmax'.
With Assumption~\ref{asm:random_variable}, the indeterminate probability (model output) is

\begin{equation}
   \label{eq:P_ipnn_Indeterminate}    
   P\left (a_{i_{j} }^{j}\mid x_{k}   \right ) := \alpha _{i_{j}}^{j}(k)
   \end{equation}

\begin{assumption}
   \label{asm:random_variable}    
   For neural networks, given an input sample $X=x_{k}$, 
   \textbf{IF} ${\textstyle \sum_{i_{j}=1}^{M_{j}}} \alpha _{i_{j}}^{j}(k) = 1$ and $\alpha _{i_{j}}^{j}(k)\in [0,1], k=1,2,\dots ,n$.
   \textbf{THEN},  
   $\left \{ a_{1}^{j} , a_{2}^{j} , \dots, a_{M_{j}}^{j} \right \}$ can be regarded as collectively exhaustive and exclusive events set, 
   they are partitions of the sample space of random variable $A^{j}, j=1,2,\dots,N$.
   \end{assumption}

According to Equation~\ref{eq:P_Y_X_via_A}, the prediction for test sample $x_{t}$ is

\begin{equation}
   \label{eq:P_Y_X_via_A_ipnn}    
   P^{\mathbb{A}} \left ( y_{l} \mid x_{t} \right ) = 
   \sum_{\mathbb{A}} 
   \left ( \frac{ \sum_{k=1}^{n}\left ( y_{l}(k)   \cdot { \prod_{j=1}^{N}}\alpha _{i_{j} }^{j}(k)  \right ) }
   {\sum_{k=1}^{n}\left ( { \prod_{j=1}^{N}}\alpha _{i_{j} }^{j}(k) \right ) } 
   \cdot\prod_{j=1}^{N}\alpha_{i_{j} }^{j}(t)  \right )
   \end{equation}

We use cross entropy as loss function:

\begin{equation}
   \label{eq:loss}
   \mathcal{L} =  -{\textstyle \sum_{l=1}^{m}} \left (  y_{l}(k)\cdot  \log P^{\mathbb{A} } \left ( y_{l} \mid x_{t} \right )\right ) 
   \end{equation}

More details on IPNN, including the introduction, related work, training Strategy, limitations, etc., can be found in Appendix~\ref{app:app_ipnn}.

\subsection{CIPNN and CIPAE}\label{sec:cipnn_and_cipae}

In \cite{cipnn}, we extended the indeterminate probability distribution to continuous random variable distribution.
We propose a general classification model called CIPNN, which works even for a 1000-dimensional latent space.

Besides, we propose a general auto-encoder called CIPAE, which do not even have the decoder component. The framework between
CIPAE and VAE~\cite{VAE} is almost the same, but VAE must use a neural network as the decoder. 
This is a special ability of our analytical solution.

\subsection{MTS forecasting}\label{sec:MTS_forecasting}

In \cite{seqip}, it shows how to consider multivariate point value as indeterminate probability distribution.
And the multivariate time series (MTS) forecasting problem is formulated as a complex distributions without relying on any neural models,
and the method even does not need any training process.
With our proposed theory, the complex distributions becomes analytical tractable, even in the presence of a 1000-dimensional latent space.

Although our proposed theory is motivated by design of new neural network architectures,
it is not limited to neural networks.
This is supported by our MTS forecasting method, which serves as strong evidence.

\section{Validations}\label{sec:experiment}

The validations in this section are focusing on our proposed axioms.
More validations or usefulness of our theory, you can also find in \cite{cipnn,seqip}.

\subsection {Evaluation on Datasets}

\begin{wrapfigure}{R}{0.6\columnwidth}    
   \centering
       \captionof{table}{Test accuracy with 3-D latent space; backbone is FCN for MNIST and Fashion-MNIST, Resnet50~\cite{resnet} for CIFAR10 and STL10.}
       \label{tab:test_accuracy}
       \begin{tabular}{cccc}
           \hline
           Dataset                    & CIPNN         & IPNN          & Simple-Softmax\\
           \hline
           MNIST                      & $95.9\pm 0.3$ & $95.8\pm 0.5$ & $97.6\pm 0.2$ \\
           \makecell{Fashion-\\MNIST} & $85.4\pm 0.3$ & $84.5\pm 1.0$ & $87.8\pm 0.2$ \\
           CIFAR10                    & $81.3\pm 1.6$ & $83.6\pm 0.5$ & $85.7\pm 0.9$ \\
           STL10                      & $92.4\pm 0.4$ & $91.6\pm 4.0$ & $94.7\pm 0.7$ \\
           \hline
       \end{tabular}   
   \end{wrapfigure}

Results on MNIST~\cite{mnist}, Fashion-MNIST~\cite{fashion_mnist}, CIFAR10~\cite{cifar10} and STL10~\cite{stl10} show that 
our proposed indeterminate probability theory is valid,
the backbone between IPNN, CIPNN and `Simple-Softmax' is the same, the last layer of the latter one is connected to softmax function. 
Although IPNN and CIPNN does not reach any SOTA, 
the results are very important evidences to our proposed mutual independence axioms, 
see Axiom~\ref{axm:A_X_independence}, Axiom~\ref{axm:Y_A_independence} and Axiom~\ref{axm:Y_X_independence}.

\subsection {Evaluation on Large Latent Space}

For IPNN, we cannot use Monte Carlo method to reduce the exponential complexity (Section~\ref{sec:Complexity}),
otherwise, IPNN will be not able to do back-propagation. Hence, we validate IPNN till to 20-D dimension.

Besides, for larger latent space, IPNN has also over-fitting problem when the dimension increases,
this is only the limitation of IPNN, not CIPNN.

\begin{table}[htbp]
   \caption{Average test accuracy of 10 times results on Large Latent Space on MNIST\@.}
   \label{tab:results_large_latent_space}
   \begin{center}
   \begin{tabular}{ccccccccc}
   \hline
   Latent space    & 5-D  & 10-D & 20-D &  50-D &  100-D &  200-D & 500-D & 1000-D \\
   IPNN            & 94.8 & 88.6 & 80.6 & - &  -  &  -  & -  & -   \\
   CIPNN           & 95.6 & 94.7 & 94.7 &  94.9 &  94.9  &  94.9  & 94.7  & 93.4 (2 times)  \\
   \hline
   \end{tabular}
   \end{center}
   \end{table}

\subsection {Evaluation with Duplicated Random Variable Inference}

If the latent variables are the same, i.e., $A^{1}$ is identical to $A^{2}$, then this is the most critical case for Axiom~\ref{axm:A_X_independence}. 

This is a critical example from Section~\ref{sec:example}.

Let $\mathbf{z} = (z,z,...)^{N}$, we use N same random variable z for the inference,
with Equation~\ref{eq:P_Y_A_2} we have

\begin{equation}
   \begin{aligned}
    P(Y=hd|z,z,...) &= \frac{\sum_{k=1}^{10}P(Y=hd|X=x_{k})\cdot P(z|X=x_{k})^{N}}{\sum_{k=1}^{10}P(z|X=x_{k})^{N}} \\
    &= \frac{\mathcal{N}(z;3,1)^{N}}{\mathcal{N}(z;3,1)^{N}+\mathcal{N}(z;-3,1)^{N}}
   \end{aligned}
\end{equation}

For next coin toss, let $P(z|X=x_{11})=\mathcal{N}(z;3,1)$, with Equation~\ref{eq:P_Y_X_via_A}, similar to Equation~\ref{eq:example_gauss_coin_toss}, we have

\begin{equation}
   P^{\mathbf{z}}(Y=hd|X=x_{11}) 
   = \frac{1}{C}\sum_{c=1}^{C}\frac{\mathcal{N}(z_{c};3,1)^{N}}{\mathcal{N}(z_{c};3,1)^{N}+\mathcal{N}(z_{c};-3,1)^{N}} \approx 1,  z_{c} \sim \mathcal{N}(z;3,1)
\end{equation}

We can see that even for duplicated random variables, our calculation results are also almost not effected.

Besides, in \cite{seqip}, we have duplicated the MTS dataset for abuse test of our theory, and results show that 
it has no negative effect to the forecasting performance.

\section{Related Work}\label{sec:related_work}

Indeterminate Probability Theory (IPT) is connected to probability foundations, uncertainty quantification, and observer-dependent frameworks. Key connections are formalized below:

\paragraph{Classical Probability Foundations} Kolmogorov's axiomatic framework~\cite{kolmogorov_probability} establishes the mathematical basis for both frequentist and Bayesian paradigms.
Frequentist approaches treat probability as long-run frequency under repeated trials, using clearly defined and unambiguously observations to infer underlying distributions.
Bayesian methods~\cite{bernardo_smith_bayesian_theory} treat probability as subjective belief, iteratively updating priors with observations to approximate reality. 
While classical frameworks model randomness in phenomena, IPT explicitly formalizes distortions from imperfect observation systems (e.g., sensor noise, cognitive biases), reducing to classical measures only when observation error vanishes (Theorem~\ref{thm:classical_subsumption}).

\paragraph{Uncertainty Modeling} Bayesian Methods, such as Bayesian Neural Networks~\cite{neal_bayes_network}, MC-Dropout~\cite{gal_mc_dropout}, and deep ensembles~\cite{lakshminarayanan_deep}, quantify model uncertainty via sampling.
Probabilistic Graphical Models (PGMs) encode conditional dependencies~\cite{pearl_bayes_network, koller_pgms} but require approximations for complex topologies. Fuzzy logic represents vagueness via membership functions~\cite{Goguen_funzzy_control}.
IPT's role provides closed-form solutions for high-dimensional $P(Y|\mathbf{z})$~\ref{sec:ip_theory} by jointly modeling system states and observer-induced distortions.

\paragraph{VAEs} Modern large-scale inference in complex probabilistic models often involves intractable posterior distributions. To address this, approximate inference techniques such as Markov Chain Monte Carlo (MCMC)~\cite{monte_carlo} and variational Bayesian methods~\cite{variational_method} have been widely adopted~\cite{VAE2}.
The Variational Autoencoder (VAE) framework provides an efficient estimator of the evidence lower bound (ELBO) for continuous latent variable models. Crucially, its encoder module functions as a stochastic observer that maps input data to parameters of an approximate posterior distribution, typically modeled as a diagonal-covariance multivariate Gaussian for simplicity~\cite{VAE}.
This diagonal covariance assumption explicitly embodies the latent dimension independence principle formalized in Axiom~\ref{axm:A_X_independence}.
Empirically, VAEs have demonstrated versatility across diverse domains including
image generation~\cite{VAE_use_image}, anomaly detection~\cite{VAE_use_anomaly} and de-noising tasks~\cite{VAE_use_denoise}~\cite{VAE_use_summary}, etc. These successful applications provide empirical support for the functional validity of Axiom~\ref{axm:A_X_independence} in practical observer implementations.

\section{Conclusion}\label{sec:conclusion}

This paper introduces \textbf{Indeterminate Probability Theory} (IPT), a novel framework for probabilistic reasoning under observation uncertainty. By explicitly modeling the interplay between ground truth and observer-dependent outputs, IPT provides a principled approach to handling discrete and continuous uncertainties within a unified formalism. The theory's conditional independence axioms (Axioms~\ref{axm:A_X_independence},\ref{axm:Y_A_independence},\ref{axm:Y_X_independence}) enable closed-form solutions for complex joint distributions, overcoming computational intractability in high-dimensional settings.

Two key applications validate IPT's efficacy:

\begin{itemize}
   \item The \textit{Indeterminate Probability Neural Network} (IPNN) enables tractable probabilistic inference in latent spaces of up to 1000 dimensions.
   \item In \textit{non-neural multivariate time series forecasting}, IPT outperforms LSTM and Transformer baselines by explicitly modeling observer-induced uncertainty.
\end{itemize}

Beyond these specific applications, IPT offers broader methodological implications across disciplines:
\begin{itemize}
\item Supervised classification may interpret data clusters as indeterminate distributions over labels.
\item Ensemble learning may formalize heterogeneous model outputs as indeterminate probabilities.
\item Physical systems may potentially benefit from IPT's observer-dependent formalism, particularly where inherent uncertainty exists (e.g., quantum measurement scenarios under Heisenberg's Uncertainty Principle \cite{uncertainty_principle}).
\end{itemize}

Notably, IPT is fully compatible with classical probability theory, subsuming it as a limiting case when observational error vanishes (see Theorem~\ref{thm:classical_subsumption}). More importantly, it provides a coherent extension for scenarios where measurements are inherently uncertain or context-dependent.

Future research directions include:
\begin{itemize}
\item Investigating theoretical connections to measure-theoretic probability and information geometry;
\item Exploring applications in causal inference and decision-making under ambiguity;
\item Conducting empirical validation in domains such as quantum measurement and nonlinear dynamical systems.
\end{itemize}

In summary, Indeterminate Probability Theory establishes a unified framework for probabilistic reasoning in contexts where observer effects cannot be neglected, offering both theoretical depth and practical utility.

\section*{Acknowledgment}

The authors would like to thank Mr. Jianlin Su for his insightful introduction to the VAE model\footnote{Available at: \url{https://kexue.fm/archives/5253}}.
The authors also gratefully acknowledge the helpful comments from anonymous reviewers of the previous submissions.

\bibliography{iclr2024_conference}
\bibliographystyle{iclr2024_conference}

\begin{appendix}

\section{An Intuitive Explanation}\label{app:intuitive_explain}

Since our proposed indeterminate probability theory is quite new, we will explain this idea by comparing it with classical probability theory, see below table:

\begin{table}[htbp]
	\caption{An intuitive comparison between classical probability theory and our proposed theory.}
	\label{tab_theory_comparison}
	\begin{center}
	\begin{tabular}{cc}
		\hline
		\makecell{Observation \\ (Classical)} & $P\left ( Y=y_{l} \mid A^{j}=a_{i_{j}}^{j} \right ) = \frac{\text{number of event }(Y=y_{l}, A^{j}=a_{i_{j}}^{j})\text{ occurs}} {\text{number of event }(A^{j}=a_{i_{j}}^{j})\text{ occurs}}$    \\
		\hline
		\makecell{Inference \\ (Classical)} & $X=x_{n+1}\xrightarrow[\textbf{Determinate}]{P\left ( A^{j}=a_{i_{j}}^{j} \mid X=x_{n+1}\right )=1}A^{j}=a_{i_{j}}^{j}\xrightarrow[\text{infer}]{P\left ( Y=y_{l} \mid A^{j}=a_{i_{j}}^{j} \right )} Y=y_{l}$   \\
		\hline
		\makecell{Observation \\ (Ours)} & $P\left ( Y=y_{l} \mid A^{j}=a_{i_{j}}^{j} \right ) = \frac{\text{sum of event }(Y=y_{l}, A^{j}=a_{i_{j}}^{j})\text{ occurs, in decimal}} {\text{sum of event }(A^{j}=a_{i_{j}}^{j})\text{ occurs, in decimal}}$      \\
		\hline
		\makecell{Inference \\ (Ours)} & $X=x_{n+1}\left \{   \begin{matrix}    \xrightarrow[]{P\left ( A^{j}=a_{1}^{j} \mid X=x_{n+1}\right )\in[0,1]} & A^{j}=a_{1}^{j} &\xrightarrow[]{P\left ( Y=y_{l} \mid A^{j}=a_{1}^{j} \right )} \\ \xrightarrow[]{P\left ( A^{j}=a_{2}^{j} \mid X=x_{n+1}\right )\in[0,1]} & A^{j}=a_{2}^{j} &\xrightarrow[]{P\left ( Y=y_{l} \mid A^{j}=a_{2}^{j} \right )}\\ \xrightarrow[]{\dots}          & A^{j}=\dots     &\xrightarrow[]{\dots}\\ \xrightarrow[\textbf{Indeterminate}]{P\left ( A^{j}=a_{M_{j}}^{j} \mid X=x_{n+1}\right )\in[0,1]} & A^{j}=a_{M_{j}}^{j} &\xrightarrow[\text{infer}]{P\left ( Y=y_{l} \mid A^{j}=a_{M_{j}}^{j} \right )}\end{matrix}\right \} Y=y_{l} $ \\
		\hline
		\multicolumn{2}{l}{Note: Replacing $A^{j}$ with joint random variable $(A^{1},A^{2},\dots,A^{N})$ is also valid for above explanation.}
	\end{tabular}
	\end{center}
\end{table}

In other word, for classical probability theory, perform a random experiment $X=x_{k}$, 
the event state is Determinate (happened or not happened), 
the probability is calculated by counting the number of occurrences, 
we define this process here as observation phase. For inference, 
perform a new random experiment $X=x_{n+1}$, the state of $A^{j}=a_{i_{j}}^{j}$ is Determinate again, 
so condition on $X=x_{n+1}$ is equivalent to condition on $A^{j}=a_{i_{j}}^{j}$, 
that may be the reason why condition on $X=x_{n+1}$ is not discussed explicitly in the past.

However, for our proposed indeterminate probability theory, 
perform a random experiment $X=x_{k}$, the event state is Indeterminate (understood as partly occurs), the probability is calculated by summing the decimal value of occurrences in observation phase. 
For inference, perform a new random experiment $X=x_{n+1}$, the state of $A^{j}=a_{i_{j}}^{j}$ is Indeterminate again, each case contributes the inference of $Y=y_{l}$, 
so the inference shall be the summation of all cases. Therefore, condition on $X=x_{n+1}$ is now different with condition on $A^{j}=a_{i_{j}}^{j}$, we need to explicitly formulate it.

Once again, our proposed indeterminate probability theory does not have any conflict with classical probability theory, the observation and inference phase of classical probability theory is one special case to our theory.

\section{Properties of Indeterminate Probability Theory}\label{app:properties}

The indeterminate probability theory may have the following properties, some have not been proved mathematically due to our limited knowledge.

\begin{proposition}
	\label{pps:pps_not_relevant}
	\textbf{IF} given $A$, $B$ and $Y$ is independent, we have $P \left ( Y \mid A, B \right ) 
	= P \left ( Y \mid A \right )$,
	\textbf{THEN}:
	\begin{equation}
	P^{(A, B)} \left ( Y\mid X = x_{n+1} \right ) 
	= P^{A} \left ( Y\mid X = x_{n+1} \right ) 
	\end{equation} 
	This property is understood as: Suppose given $A$, $B$ and $Y$ is independent, so $B$ does not contribute for the inference.
	\end{proposition}

\begin{proof}

\begin{equation}
	\begin{aligned}
	& P^{(A, B)} \left ( Y\mid X = x_{n+1} \right ) \\
	& = \sum_{A,B} \left ( P\left ( Y\mid A,B \right ) \cdot P\left ( A,B \mid X = x_{n+1} \right ) \right ) \\
	& = \sum_{A,B} \left ( P\left ( Y\mid A \right ) \cdot P\left ( A \mid X = x_{n+1} \right ) \cdot P\left ( B \mid X = x_{n+1} \right ) \right )\\
	& = \sum_{A} \left ( P\left ( Y\mid A \right ) \cdot P\left ( A \mid X = x_{n+1} \right )\right )
	\cdot \sum_{B} P\left ( B \mid X = x_{n+1} \right ) \\	
	& = \sum_{A} \left ( P\left ( Y\mid A \right ) \cdot P\left ( A \mid X = x_{n+1} \right )\right ) \\
	& = P^{A} \left ( Y\mid X = x_{n+1} \right )
	\end{aligned}
	\end{equation} 
 
\end{proof}

\begin{hypothesis}
    \label{hyp:hyp_independent}
	\textbf{IF} given $A$, $Y$ and $V$ is independent,
	\textbf{THEN}:
	\begin{equation}
	P^{A} \left ( Y, V\mid X = x_{n+1} \right ) 
	= P^{A } \left ( Y\mid X = x_{n+1} \right ) \cdot P^{A } \left (V\mid X = x_{n+1} \right )
	\end{equation}  
	This property is understood as: Given $A$, $ Y$ and  $ V$is independent, so the inference outcome is also independent.
	\end{hypothesis}


		

\begin{hypothesis}
	\label{hyp:hyp_recurse_Y}
	Let $P\left (A \mid X = x_{n+1} \right ) \in [ 0,1 ) $ and 
	\begin{equation}
	\begin{aligned}
	& P \left ( Y^{0} = y_{l}\mid X = x_{n+1} \right )  = P^{A} \left ( Y = y_{l}\mid X = x_{n+1} \right ) \\
	& P \left ( Y^{1} = y_{l}\mid X = x_{n+1} \right )  = P^{Y^{0}} \left ( Y = y_{l}\mid X = x_{n+1} \right ) \\
	& P \left ( Y^{2} = y_{l}\mid X = x_{n+1} \right )  = P^{Y^{1}} \left ( Y = y_{l}\mid X = x_{n+1} \right ) \\
	& \dots
	\end{aligned}
	\end{equation}  
	Our hypothesis is:
	\begin{equation}
	P^{Y^{\infty}} \left ( Y = y_{l}\mid X = x_{n+1} \right )
	= \frac{1}{m}, l = 1,2,\dots, m.
	\end{equation}  
	This property is understood as: The inference accuracy will become poor as the information is transmitted one after another (from $Y^{i-1}$ to $Y^{i}$).
	\end{hypothesis}

\begin{hypothesis}
	\label{hyp:hyp_multi_A}
	Let  $P\left (Y = y_{l} \mid X = x_{n+1} \right ) \in \{ 0,1 \}$ and $P\left (A \mid X = x_{n+1} \right ) \in [ 0,1 ) $.
	Our hypothesis is:
	\begin{equation}
	\max_{l=1,2,\dots,m} P^{(A,A)} \left ( Y = y_{l}\mid X = x_{n+1} \right ) 
	> \max_{l=1,2,\dots,m} P^{(A)} \left ( Y = y_{l}\mid X = x_{n+1} \right ) 
	\end{equation}  
	This property is understood as: The inference tendency will get more stronger with more same information ($A,A$).
	\end{hypothesis}

\section{Why is Indeterminate Probability Theory is Good?}\label{app:app_advantage}

\begin{table}[htbp]
	\caption{Comparison of independence assumptions}
	\label{tab:Comparison_Independence}
	\begin{center}
	\begin{tabular}{cccc}
		\hline
      & Assumption & Validity & Assumption Range \\
		\hline
      Example & $A^{1},\dots,A^{N}$ independent &  Strongest assumption  & all samples \\
      Naïve Bayes & Given $Y$, $A^{1},\dots,A^{N}$ independent & Strong assumption & few samples \\
      Ours & See our Candidate Axioms. & No exception  &  one sample \\ 
		\hline
	\end{tabular}
	\end{center}
\end{table}

Let's think the independent assumption in another way. Sometimes, $A^{1},A^{2},\dots,A^{N}$ independence assumption is strong.  
Nevertheless, in the case of Naïve Bayes, the whole samples are partitioned into small groups due to condition on $Y=y_{l}$, 
the conditional independence maybe not strong anymore. This maybe the reason why Naïve Bayes is successful for many applications.

For our proposed Candidate Axioms, the whole samples are partitioned into a single sample due to $X=x_{k}$, our assumptions are the most weak one.
For example, even if $A^1$ is identical to $A^2$, our independent assumptions still hold true. 
Furthermore, we have already conducted tests with thousand of latent variables in CIPNN, these assumptions have proven to remain valid. 
In IPNN, you can test with a few variables due to the exponentially large space size during the training phase, but not during the prediction phase (Monte Carlo).

\section{Comparison}\label{app:comparison}

\textbf{General frequency-based Probability Form} 
\begin{itemize}
   \item \textbf{Equation}:
   \begin{equation}
   \label{eq:general_probability}
   \frac{\text{number of event }(Y=y_{l}, A^{1}=a_{i_{1}}^{1},\dots, A^{N}=a_{i_{N}}^{N})\text{ occurs}} {\text{number of event }(A^{1}=a_{i_{1}}^{1},\dots, A^{N}=a_{i_{N}}^{N})\text{ occurs}}
   \end{equation}
   \item \textbf{Assumption}: No assumption.
   \item \textbf{Limitations}: \\1. Not applicable if $A^{j}$ is continuous. \\2. Not applicable for indeterminate case. \\3. Joint sample space is exponentially large. 
   \item \textbf{Space Size}: $m\cdot\prod_{j=1}^{N}M_{j}$
\end{itemize}

\textbf{Naïve Bayes Form} 
\begin{itemize}
   \item \textbf{Equation}:
   \begin{equation}
   \label{eq:naive_bayes}
   \frac{P(Y=y_{l})\cdot\prod_{j=1}^{N}P(A^{j}=a_{i_{j}}^{j}\mid Y=y_{l})}{P(A^{1}=a_{i_{1}}^{1},\dots, A^{N}=a_{i_{N}}^{N})}
   \end{equation}
   \item \textbf{Assumption}: Given $Y$, $A^{1},A^{2},\dots,A^{N}$ conditionally independent.
   \item \textbf{Limitations}: \\
   1. Assumption is strong. \\
   2. $P(A^{j}=a_{i_{j}}^{j}\mid Y=y_{l})$ is not always solvable.
   \item \textbf{Space Size}: $m\cdot\sum_{j=1}^{N}M_{j}$
\end{itemize}

\textbf{Indeterminate Probability Form} 
\begin{itemize}
   \item \textbf{Equation}: 
   Equation~\ref{eq:P_Y_A_2}
   \item \textbf{Assumption}: Given $X$, $A^{1},A^{2},\dots,A^{N}$ and $Y$ conditionally independent. see Axiom~\ref{axm:A_X_independence} and Axiom~\ref{axm:Y_A_independence}.
   \item \textbf{Limitations}: No.
   (Joint sample space is exponentially large only when Monte Carlo method is not used.) 
   \item \textbf{Space Size}: $m\cdot n\cdot N\cdot C$ (or $m\cdot\prod_{j=1}^{N}M_{j}$ without Monte Carlo method, see Section~\ref{sec:Complexity}.)
\end{itemize}

Due to the limitations of general probability form and Naïve Bayes form, MCMC~\cite{monte_carlo} and variational inference methods~\cite{variational_method}
as approximate solutions are well developed in the past.

\section{IPNN}\label{app:app_ipnn}

\subsection{Introduction}\label{sec:app_intro}

 Humans can distinguish at least 30,000 basic object categories~\cite{human_Recognition}, classification of all these 
 would have two challenges: It requires huge well-labeled images; Model with softmax for large scaled datasets is computationally expensive.
 Zero-Shot Learning \textendash\enspace ZSL~\cite{ZSL,ZSL_Summary} method provides an idea for solving the first problem, which is an attribute-based classification method. 
 ZSL performs object detection
 based on a human-specified high-level description of the target object instead of training images, like shape, color or even geographic information.
 But labelling of attributes still needs great efforts and expert experience. Hierarchical softmax can solve the computationally expensive problem,
 but the performance degrades as the number of classes increase~\cite{softmax}.
 
 Probability theory has not only achieved great successes in the classical area, 
 such as Na\"{\i}ve Bayesian method~\cite{bayes_study}, 
 but also in deep neural networks (VAE~\cite{VAE}, ZSL, etc.) over the last years.  
 However, both have their shortages: Classical probability can not extract features from samples; 
 For neural networks, the extracted features are usually abstract and cannot be directly used for numerical probability calculation.
 What if we combine them?
 
 There are already some combinations of neural network and bayesian approach, 
 such as probability distribution recognition~\cite{bayes_combine_recognition,bayes_combine_timeseries},
 Bayesian approach are used to improve the accuracy of neural modeling~\cite{bayes_model}, \etc. However, current combinations
 do not take advantages of ZSL method.
 
We propose an approach to solve the mentioned problems, 
and we propose a novel unified combination of (indeterminate) probability theory and deep neural network.
The neural network is used to extract attributes which are defined as discrete random variables, and the inference model for classification task is derived.
Besides, these attributes do not need to be labeled in advance.

 \subsection{Related Work}\label{app:Related_work}
 
 \paragraph{Tractable Probabilistic Models.} There are a large family of tractable models including probabilistic circuits~\cite{PCs_Unifying,sparse_PCs},
 arithmetic circuits~\cite{arithmetic_factor_belief,learn_arithmetic}, sum-product networks~\cite{sum_product}, cutset networks~\cite{cutset}, and-or search spaces~\cite{and_or}, and probabilistic sentential decision diagrams~\cite{decision_diag}.
 The analytical solution of a probability calculation is defined as occurrence, $P(A=a) = \frac{\text{number of event }(A=a)\text{ occurs}}{\text{number of random experiments}} $,
 which is however not focused in these models. Our proposed IPNN is fully based on event occurrence and is an analytical solution. 
 
 \paragraph{Deep Latent Variable Models.} DLVMs are probabilistic models and can refer to the use of neural networks to perform
 latent variable inference~\cite{DLVMs_Tutorial}. Currently, the posterior calculation of continuous latent variables is regarded as intractable~\cite{VAE_2019}, 
 VAEs~\cite{VAE,VAE2,VAE3,VAE_binary} use variational inference method~\cite{variational_method} as approximate solutions. 
 Our proposed IPNN is one DLVM with discrete latent variables and the intractable posterior calculation is now analytically solved with our proposed theory.

\subsection{Training}\label{app:training}

\subsubsection{Training Strategy}

Given an input sample $x_{t}$ from a mini batch, with a minor modification of Equation~\ref{eq:P_Y_X_via_A_ipnn}:

\begin{equation}
   \label{eq:P_Y_X_via_A_Train}
   P^{\mathbb{A} } \left ( y_{l} \mid x_{t} \right ) 
   \approx \sum_{\mathbb{A}}
   \left ( \frac{\max(H+h(\bar{t} ),\epsilon)}
   {\max(G+g(\bar{t} ),\epsilon)} \cdot { \prod_{j=1}^{N}}\alpha _{i_{j} }^{j}(t) \right )
   \end{equation}

\begin{align} 
   \label{eq:hh}
   &h(\bar{t} )={\textstyle \sum_{k=b\cdot (\bar{t} -1)+1}^{b\cdot \bar{t} }}\left (  y_{l}(k)\cdot  
   {\textstyle  \prod_{j=1}^{N}}\alpha _{i_{j} }^{j}(k) \right )\\
   \label{eq:gg}
   &g(\bar{t} )= {\textstyle \sum_{k=b\cdot (\bar{t} -1)+1}^{b\cdot \bar{t} }}
   \left (  {\textstyle  \prod_{j=1}^{N}}\alpha _{i_{j} }^{j}(k) \right )\\
   \label{eq:H}
   &H= {\textstyle \sum_{k=\max(1,\bar{t} -T)}^{\bar{t} -1}} h(k),\text{for } \bar{t} =2,3,\dots\\
   \label{eq:G}
   &G= {\textstyle \sum_{k=\max(1,\bar{t} -T)}^{\bar{t} -1}} g(k),\text{for } \bar{t} =2,3,\dots
\end{align}

Where $b$ is for batch size, $\bar{t}  = \left \lceil \frac{t}{b}  \right \rceil , t = 1,2,\dots,n$. 
Hyper-parameter T is for forgetting use, i.e., $H$ and $G$ are calculated from the recent T batches. Hyper-parameter T is introduced because at 
beginning of training phase the calculated result with Equation~\ref{eq:P_Y_A_2} is not good yet. 
And the $\epsilon$ on the denominator is to avoid dividing zero, 
the $\epsilon$ on the numerator is to have an initial value of 1. 
Besides, $H$ and $G$ are not needed for gradient updating during back-propagation.  
The detailed algorithm implementation is shown in Algorithm~\ref{alg:train}.

\begin{algorithm}[H]
	\captionof{algorithm}{IPNN training}
	\label{alg:train}
	\textbf{Input}: A sample $x_{t}$ from mini-batch \\
	\textbf{Parameter}: Split shape, forget number $T$, $\epsilon$, learning rate $\eta$.\\
	\textbf{Output}: Posterior $P^{\mathbb{A} } \left ( y_{l} \mid x_{t} \right )$ 
	\begin{algorithmic}[1]
		\STATE Declare default variables: $H, G, hList, gList$
		\FOR{$\bar{t}=1,2,\dots$ Until Convergence}
		\STATE Compute $h, g$ with Equation~\ref{eq:hh} and Equation~\ref{eq:gg}
		\STATE Record: $hList.append(h), gList.append(g)$
		\IF{$\bar{t}>T$}
		\STATE Forget: $H = H - hList[0], G = G - gList[0]$
		\STATE Remove first element from $hList, gList$
		\ENDIF
		\STATE Compute posterior with Equation~\ref{eq:P_Y_X_via_A_Train}: $P^{\mathbb{A} } \left ( y_{l} \mid x_{t} \right )$ 
		\STATE Compute loss with Equation~\ref{eq:loss}: $\mathcal{L}(\theta )$
		\STATE Update model parameter: $\theta = \theta - \eta\nabla \mathcal{L}(\theta )$
		\STATE Update for next loop: $H=H+h, G=G+g$
		\ENDFOR
		\STATE \textbf{return} model and the probability
	\end{algorithmic}
\end{algorithm}

With Equation~\ref{eq:P_Y_X_via_A_Train} we can get that $P^{\mathbb{A} } \left ( y_{l} \mid x_{1} \right )=1$ 
for the first input sample if $y_{l}$ is the ground truth and batch size is 1. 
Therefore, for IPNN the loss may increase at the beginning and fall back again while training.

\subsubsection{Multi-degree Classification (Optional)}\label{app:classification}

In IPNN, the model outputs N different random variables $A^{1},A^{2},\dots,A^{N}$, 
if we use part of them to form sub-joint sample spaces, we are able of doing sub classification task, 
the sub-joint spaces are defined as $\Lambda^{1}\subset \mathbb{A} ,\Lambda^{2}\subset \mathbb{A} ,\dots$ 
The number of sub-joint sample spaces is:

\begin{equation}
   \label{eq:sub_joint_spaces}
   \sum_{j=1}^{N}\binom{N}{j} 
   =\sum_{j=1}^{N}\left ( \frac{N!}{j!(N-j)!}  \right ) 
   \end{equation}

If the input samples are additionally labeled for part of sub-joint sample spaces\footnote{It is labelling of input samples, not sub-joint sample points.},
defined as $Y^{\tau} \in \{y_{1}^{\tau},y_{2}^{\tau},\dots,y_{m^{\tau}}^{\tau} \}$.
The sub classification task can be represented as $\left \langle X,\Lambda^{1},Y^{1} \right \rangle ,\left \langle X,\Lambda^{2},Y^{2} \right \rangle,\dots$
With Equation~\ref{eq:loss} we have,

\begin{equation}
   \label{eq:sub_losses}
   \mathcal{L}^{\tau} = -{\textstyle \sum_{l=1}^{m^{\tau}}} \left (    y_{l}^{\tau}(k)\cdot  
   \log P^{\Lambda^{\tau} } \left ( y_{l}^{\tau} \mid x_{t} \right ) \right ), \tau = 1,2,\dots
   \end{equation}

Together with the main loss, the overall loss is $\mathcal{L}+\mathcal{L}^{1}+\mathcal{L}^{2}+\dots$
In this way, we can perform multi-degree classification task. 
The additional labels can guide the convergence of the joint sample spaces and speed up the training process, as discussed later in Appendix~\ref{app:avoiding_local_minimum}. 

\subsubsection{Multi-degree Unsupervised Clustering}\label{app:cluster}

If there are no additional labels for the sub-joint sample spaces, the model are actually doing
unsupervised clustering while training.
And every sub-joint sample space describes one kind of clustering result, 
we have Equation~\ref{eq:sub_joint_spaces} number of clustering situations in total.

\subsubsection{Designation of Joint Sample Space}\label{app:split_shape}

As in Appendix~\ref{app:glb_convergence} proved, we have following proposition:

\begin{proposition}
   \label{pps:convergence}
   For $P(y_{l}|x_{k})=y_{l}(k) \in \{0,1\}$ hard label case, 
   IPNN converges to global minimum only when $P\left (y_{l}| a_{i_{1}}^{1},a_{i_{2}}^{2},\dots,a_{i_{N}}^{N} \right ) = 1,
   \text{ for } \prod_{j=1}^{N}\alpha_{i_{j} }^{j}(t)  > 0, i_{j} = 1,2,\dots,M_{j}$. In other word, each joint sample point corresponds to an unique category.
   However, a category can correspond to one or more joint sample points. 
\end{proposition}

\begin{corollary}
   \label{cor:condition}
   The necessary condition of achieving the global minimum is when the split shape defined in Equation~\ref{eq:split_shape} satisfies: 
   ${\textstyle \prod_{j=1}^{N}}M_{j} \ge m$, where $m$ is the number of classes. That is, for a classification task, 
   the number of all joint sample points is greater than the classification classes.
\end{corollary}

Theoretically, if model with 100 output nodes are split into 10 equal parts, it can classify 10 billion categories, validation result see Appendix~\ref{app:avoiding_local_minimum}.
Besides, the unsupervised clustering (Appendix~\ref{app:cluster}) depends on the input sample distributions, the split shape shall not violate
from multi-degree clustering. For example, if the main attributes of one dataset shows three different colors, and your split shape is $\{2,2,\dots\}$,
this will hinder the unsupervised clustering, in this case, the shape of one random variable is better set to 3. And as in Appendix~\ref{app:local_convergence} also analyzed, 
there are two local minimum situations, improper split shape will make IPNN go to local minimum.

In addition, the latter part from Proposition~\ref{pps:convergence} also implies that IPNN may be able of doing further unsupervised classification task, this is beyond the scope of this discussion.

\subsection{Results of IPNN}

\subsubsection{Unsupervised Clustering}

\begin{figure}[htbp]
\centering
\begin{minipage}[b]{0.49\columnwidth}
      \centering
      \includegraphics[width=0.85\columnwidth]{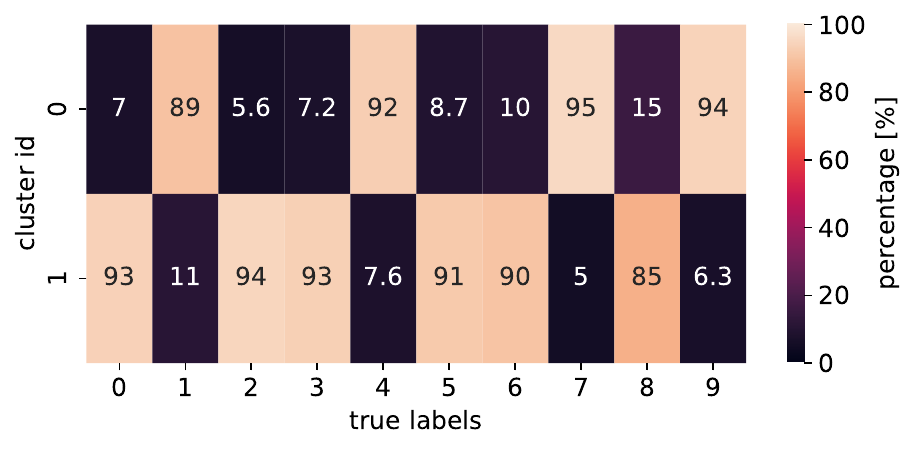}
\end{minipage}
\hfill
\begin{minipage}[b]{0.5\columnwidth}              
   \centering
   \begin{equation} \nonumber
      \begin{aligned}
      &\text{percentage} = \\
      &\frac{1}{\text{round}}  \cdot \sum_{i=1}^{\text{round}} \frac{
      \begin{matrix}
      \text{ number of samples with label } l \\
      \text{ in one cluster at }  i^{th} \text{ round}
      \end{matrix}}
      {\text{number of samples with label }l} 
      \end{aligned}
      \end{equation}
   \vspace{1em}    
\end{minipage}        
\caption{Unsupervised clustering results on MNIST\@: test accuracy $95.1\pm 0.4$, $\epsilon = 2$, batch size $b=64$, forget number $T=5$, epoch is 5 per round.
The test was repeated for 876 rounds with same configuration (different random seeds) in order to check the stability of clustering performance,
each round clustering result is aligned using Jaccard similarity~\cite{jaccard}.}
\label{fig:cluster_mnist}
\end{figure}

As in Appendix~\ref{app:cluster} discussed, IPNN is able of performing unsupervised clustering, we evaluate it on MNIST\@.
The split shape is set to $\{2,10\}$, it means we have two random variables, and the first random variable is used to divide
MNIST labels $0,1,\dots 9$ into two clusters.
The cluster results is shown in Figure~\ref{fig:cluster_mnist}.

We find only when $\epsilon$ in Equation~\ref{eq:P_Y_X_via_A_Train} is set to a relative high value that IPNN prefers to put number 1,4,7,9 into one cluster and the rest into another cluster, 
otherwise, the clustering results is always different for each round training.
The reason is unknown, our intuition is that high $\epsilon$ makes that each category catch the free joint sample point more harder, categories have similar
attributes together will be more possible to catch the free joint sample point.

\subsubsection{Hyper-parameter Analysis}

\begin{figure}[htbp]
\centering
      \begin{subfigure}[b]{0.3\columnwidth}
            \centering
            \includegraphics[width=1\columnwidth]{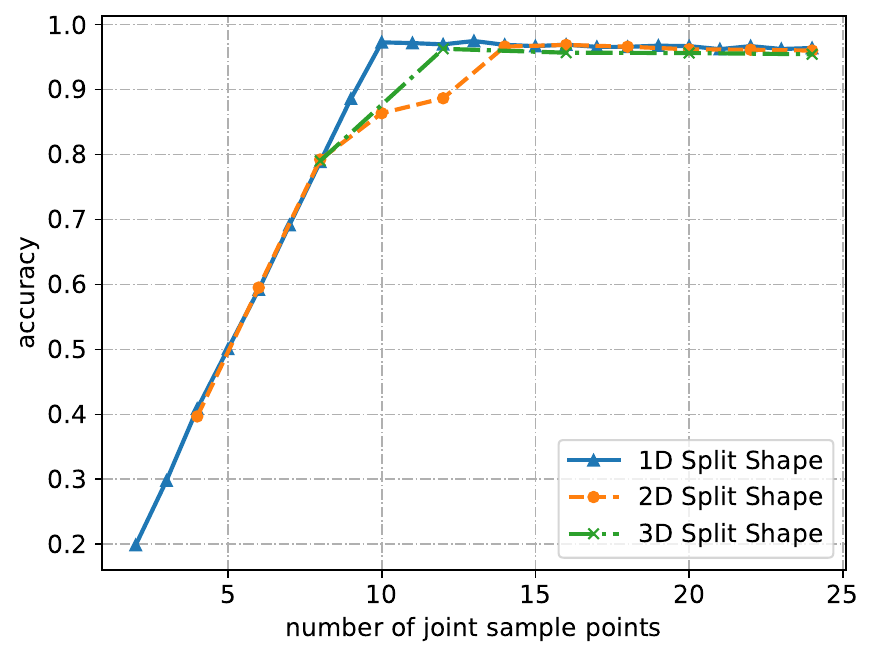}
            \caption{split shape}
            \label{fig:split_shape}
      \end{subfigure}
      \begin{subfigure}[b]{0.3\columnwidth}
            \centering
            \includegraphics[width=1\columnwidth]{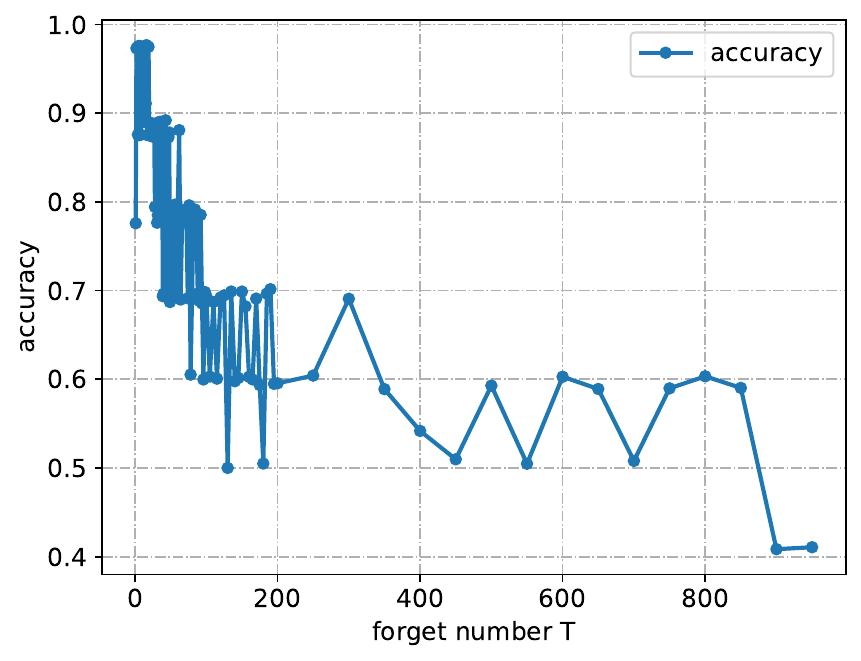}
            \caption{forget number T}
            \label{fig:T}
      \end{subfigure}
      \caption{(a) Impact Analysis of split shape with MNIST\@:
      1D split shape is for $\{\tau\}, \tau = 2,3,\dots, 24$.
      2D split shape is for $\{2,\tau\}, \tau = 2,3,\dots, 12$.
      3D split shape is for $\{2,2,\tau\}, \tau = 2,3,\dots, 6$.
      The x-axis is the number of joint sample points calculated with ${\textstyle \prod_{j=1}^{N}}M_{j}$, see Equation~\ref{eq:split_shape}. \\
      (b) Impact Analysis of forget number T with MNIST\@: Split shape is  $\{10\}$.}
      \label{fig:hyperparamers}
\end{figure}

IPNN has two import hyper-parameters: split shape and forget number T. In this section, we have analyzed it with test on MNIST, batch size is set to 64, $\epsilon = 10^{-6}$.
As shown in Figure~\ref{fig:split_shape}, if the number of joint sample points is smaller than 10, IPNN is not able of making
a full classification and its test accuracy is proportional to number of joint sample points, as number of joint sample points increases over 10,
IPNN goes to global minimum for both 3 cases, this result is consistent with our analysis. However, we have exceptions, the accuracy of split shape with $\{2, 5\}$ and $\{2, 6\}$ is not high. From Figure~\ref{fig:cluster_mnist}
we know that for the first random variable, IPNN sometimes tends to put number 1,4,7,9 into one cluster and the rest into another cluster, 
so this cluster result request that the split shape need to be set minimums to $\{2, \ge 6\}$ in order to have enough free joint sample points.
That's why the accuracy of split shape with $\{2, 5\}$ is not high. (For $\{2, 6\}$ case, only three numbers are in one cluster.)

Another test in Figure~\ref{fig:T} shows that IPNN will go to local minimum as forget number T increases and cannot go to global minimum without further actions,
hence, a relative small forget number T shall be found with try and error.

\subsection{Conclusion}
For a classification task, we proposed an approach to extract the attributes of input samples as random variables,
and these variables are used to form a large joint sample space. After IPNN converges to global minimum, each joint sample point will correspond
to an unique category, as discussed in Proposition~\ref{pps:convergence}. As the joint sample space increases exponentially, the classification capability
of IPNN will increase accordingly.

We can then use the advantages of classical probability theory, for example,
for very large joint sample space, we can use the Bayesian network approach or mutual independence among variables
(see Appendix~\ref{app:independent}) to simplify the model 
and improve the inference efficiency, in this way,
a more complex Bayesian network could be built for more complex reasoning task.

\subsection{Global Minimum Analysis}\label{app:glb_convergence}

\begin{proof}[Proof of Proposition~\ref{pps:convergence}]

	Equation~\ref{eq:P_Y_X_via_A_ipnn} can be rewritten as:

\begin{equation}
	\label{eq:convergence_p}
	P^{\mathbb{A} } \left ( y_{l} \mid x_{t} \right )
	=\sum_{\mathbb{A}} 
	\left (p_{\mathbb{A}}\cdot   {\textstyle \prod_{j=1}^{N}\alpha_{i_{j} }^{j}(t) }  \right ) 
	\end{equation}

Where,

\begin{equation}
	\label{eq:p_Y_A_shorted}
	p_{\mathbb{A}} = P\left ( y_{l} \mid a_{i_{1}}^{1} , a_{i_{2}}^{2} , \dots, a_{i_{N}}^{N} \right ) 
	\end{equation}

Theoretically, for $P(y_{l}|x_{k})=y_{l}(k) \in \{0,1\}$ hard label case, model converges to global minimum when the train and test loss is zero~\cite{convergence_paper}, 
and for the ground truth $y_{l}(t)=1$, with Equation~\ref{eq:loss} we have:

\begin{equation}
	\label{eq:p_equal_one}
	\sum_{\mathbb{A}} 
	\left (p_{\mathbb{A}}\cdot   {\textstyle \prod_{j=1}^{N}\alpha_{i_{j} }^{j}(t) }  \right )  = 1
	\end{equation}

Subtract the above equation from Equation~\ref{eq:Sum_P_A_X} gives:

\begin{equation}
	\label{eq:conv_eq:zero}
	\sum_{\mathbb{A}} 
	\left ( (1-p_{\mathbb{A}}) \cdot \prod_{j=1}^{N}\alpha_{i_{j} }^{j}(t) \right )  = 0
	\end{equation}

Because $\prod_{j=1}^{N}\alpha_{i_{j} }^{j}(t) \in [0,1]$ and $(1-p_{\mathbb{A}}) \in [0,1]$, 
The above equation is then equivalent to:

\begin{equation}
\label{eq:convergence_condition}
	p_{\mathbb{A}}  = 1, 
	\text{ for } \prod_{j=1}^{N}\alpha_{i_{j} }^{j}(t)  > 0, i_{j} = 1,2,\dots,M_{j}.
	\end{equation}

\end{proof}

\subsection{Local Minimum Analysis}\label{app:local_convergence}

Equation~\ref{eq:convergence_p} can be further rewritten as:

\begin{equation}
	\label{eq:local_p}
	P^{\mathbb{A} } \left ( y_{l} \mid x_{t} \right ) 
	= \sum_{i_{\tau}=1}^{M_{\tau}}
	\left ( \alpha_{i_{\tau} }^{\tau}(t) \cdot \sum_{\Lambda} 
	\left ( p_{\mathbb{A}}\cdot   {\textstyle \prod_{j=1,j \ne \tau}^{N}\alpha_{i_{j} }^{j}(t)}   \right ) \right )  \\
	= {\textstyle \sum_{i_{\tau}=1}^{M_{\tau}}
	\left ( \alpha_{i_{\tau} }^{\tau}(t) \cdot p_{i_{\tau}} \right )}
\end{equation}

Where $\Lambda =(A^{1},\dots,A^{j},\dots,A^{N}) \subset \mathbb{A}, j \ne \tau$ and,

\begin{equation}
	p_{i_{\tau}} = {\sum_{\Lambda} \left (p_{\mathbb{A}}\cdot  {\textstyle \prod_{j=1,j \ne \tau}^{N}
	\alpha_{i_{j} }^{j}(t)  } \right ) }
	\end{equation}

Substitute Equation~\ref{eq:local_p} into Equation~\ref{eq:loss}, and for the ground truth
$y_{l}(t)=1$ the loss function can be written as:
	
\begin{equation}
	\label{eq:loss_for_analysis}
	\mathcal{L} = - \log({\textstyle \sum_{i_{\tau}=1}^{M_\tau}
		\left ( \alpha_{i_{\tau} }^{\tau}(t) \cdot p_{i_{\tau}} \right ) } )
	\end{equation}

Let the model output before softmax function be $z_{i_{j}}$, we have:

\begin{equation}
	\alpha_{i_{\tau}}^{\tau}(t) = 
	\frac{e^{z_{i_{\tau}}}}{ {\textstyle \sum_{i_{j}=1}^{M_{j}}e^{z_{i_{j}}}} } 
	\end{equation}

In order to simplify the calculation, we assume $p_{\mathbb{A}}$ defined in Equation~\ref{eq:p_Y_A_shorted} is constant during back-propagation.
so the gradient is:

\begin{equation}
	\label{eq:convergence_grad}
	\frac{\partial \mathcal{L}}{\partial z_{i_{\tau}}} 
	= - \frac{\alpha_{i_{\tau}}^{\tau}(t) \cdot {\textstyle \sum_{i_{j}=1,i_{j} \ne i_{\tau}}^{M_{j}}} \left ( e^{z_{i_{j}}}\cdot(p_{i_{\tau}}-p_{i_{j}}) \right )  }
	{\sum_{i_{j}=1}^{M_j} \left ( e^{z_{i_{j}}}\cdot p_{i_{j}} \right )  } 
	\end{equation}

Therefore, we have two kind of situations
that the algorithm will go to local minimum:

\begin{equation}
	\frac{\partial \mathcal{L}}{\partial z_{i_{\tau}}} 
	=\begin{cases}
	\to 0,&\text{if } \left |  z_{i_{\tau}}-z_{i_{j}} \right | \to \infty \\
	0,&\text{if } p_{i_{\tau}}=p_{i_{j}}\\
	Nonezero, &o.w.
	\end{cases}
	\end{equation}

Where $i_{\tau} = 1,2,\dots,M_{\tau}$.

The first local minimum usually happens when Corollary~\ref{cor:condition} is not satisfied,
that is, the number of joint sample points is smaller than the classification classes, the results are shown in Figure~\ref{fig:split_shape}.

If the model weights are initialized to a very small value, the second local minimum may happen at the beginning of training.
In such case, all the model output values are also small which will result in $\alpha_{1}^{j}(t) \approx \alpha_{2}^{j}(t) \approx  \dots \approx \alpha_{M_{j}}^{j}(t)$,
and it will further lead to all the $p_{i_{\tau}}$ be similar among each other.
Therefore, if the model loss reduces slowly at the beginning of training, the model weights is suggested to be initialized to an relative high value.
But the model weights shall not be set to too high values, otherwise it will lead to first local minimum. 

\begin{figure}[htbp]
	\vskip 0.2in
	\begin{center}
		\centerline{\includegraphics[width=1\columnwidth]{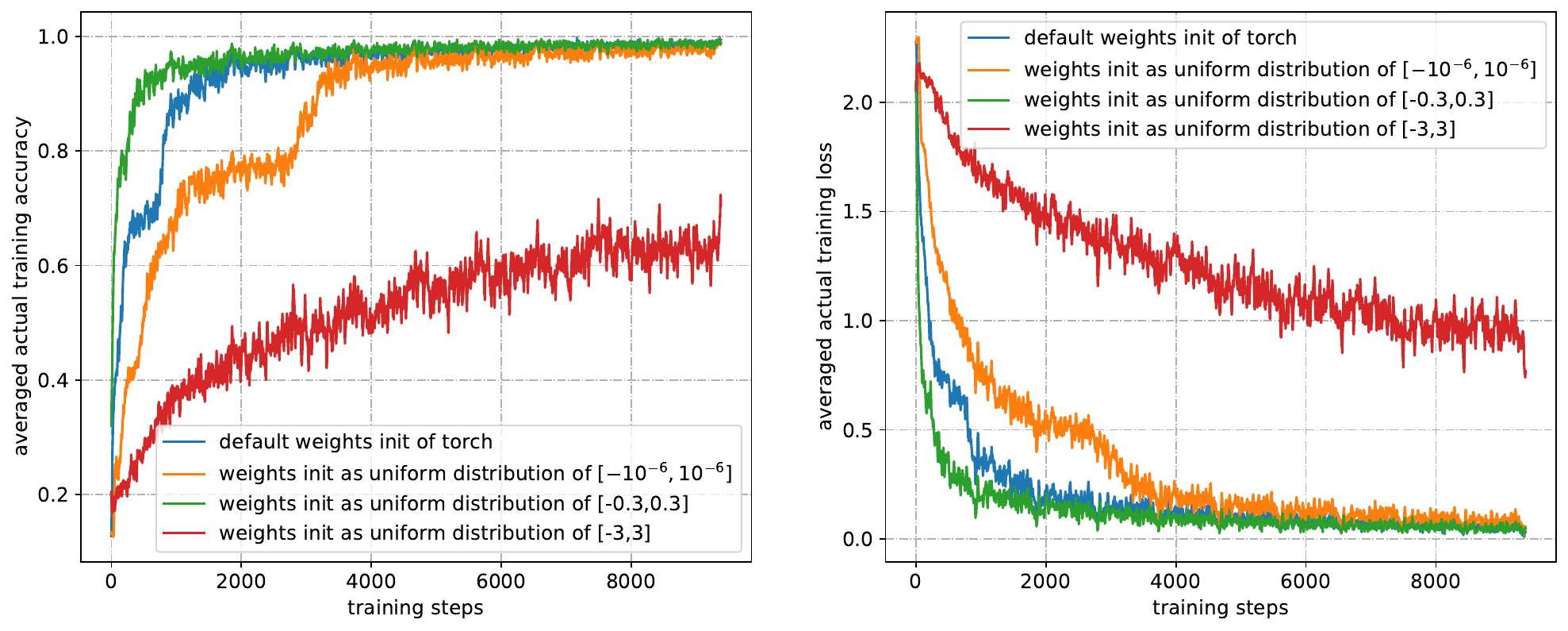}}
		\caption{Model weights initialization impact analysis on MNIST\@. 
		Split shape is $\{2,10\}$, batch size is 64, forget number $T=5,\enspace \epsilon=10^{-6}$.}
		\label{fig:weights_init}
	\end{center}
	\vskip -0.2in
\end{figure}

As shown in Figure~\ref{fig:weights_init}, if model weights are initialized to uniform distribution of $\left [ -10^{-6},10^{-6} \right ] $,
its convergence speed is slower than the model weights initialized to uniform distribution of $\left [ -0.3,0.3\right ]$. 
Besides, model weights initialized to uniform distribution of $\left [ -3,3\right ] $ get almost stuck at local minimum and cannot go to global minimum.
This result is consistent with our analysis.

\subsubsection{Avoiding Local Minimum with Multi-degree Classification} \label{app:avoiding_local_minimum}

Another experiment is designed by us to check the performance of multi-degree classification (see Appendix~\ref{app:classification}): classification of binary vector into decimal value.
The binary vector is the model inputs from `000000000000' to `111111111111', which are labeled from 0 to 4095.
The split shape is set to $\{M_{1}=2,M_{2}=2,\dots, M_{12}=2\}$, which is exactly able of making a full classification.  
Besides, model weights are initialized as uniform distribution of $[-0.3,0.3]$, as discussed in Appendix~\ref{app:local_convergence}.  

\begin{figure}[htbp]
	\centering
	\includegraphics[width=0.5\columnwidth]{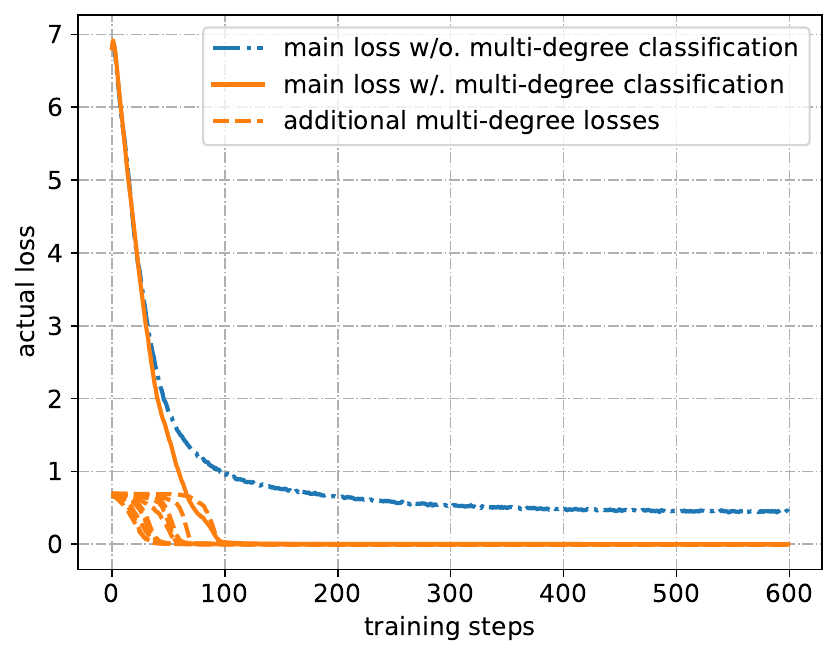}
	\caption{Loss of multi-degree classification of `binary to decimal' on train dataset. 
			Input samples are additionally labeled with $Y^{i} \in \{0,1\}$ for $i^{th}$ bit is 0 or 1, respectively. 
			$Y^{i}$ corresponds to sub-joint sample space $\Lambda^{i}$ with split shape $\{M_{i}=2\}, i = 1,2,\dots12$.  
			Batch size is 4096, forget number $T=5,\enspace \epsilon=10^{-6}$.}
	  \label{fig:multi_classification}
  \end{figure}

The result is shown in Figure~\ref{fig:multi_classification}, IPNN without multi degree
classification goes to local minimum with only $69.5\%$ train accuracy.
We have only additionally labeled for 12 sub-joint spaces, and IPNN goes to global minimum with $100\%$ train accuracy.

Therefore, with only ${\textstyle \sum_{1}^{12}} 2=24$ output nodes, IPNN can classify 4096 categories.
Theoretically, if model with 100 output nodes are split into 10 equal parts, it can classify 10 billion categories.
Hence, compared with the classification model with only one `softmax' function, IPNN has no computationally expensive problems (see Section~\ref{sec:introduction}).

\subsection{Mutual Independency}\label{app:independent}

If we want the random variables $A^{1},A^{2},\dots,A^{N}$ partly or fully mutually independent, 
we can use their mutual information as loss function:

\begin{gather}
	\label{eq:loss_independent}
	\mathcal{L}^{*} = KL\left ( P(A^{1},A^{2},\dots,A^{N}), \prod_{j=1}^{N}P(A^{j})  \right ) 
	=\sum_{\mathbb{A}}\left (  P\left (a_{i_{1}}^{1} , \dots, a_{i_{N}}^{N} \right ) \cdot
	\log \frac{P\left (a_{i_{1}}^{1} , \dots, a_{i_{N}}^{N} \right )}{\prod_{j=1}^{N}P(a_{i_{j}}^{j})}  \right ) \\ \nonumber
	= \sum_{\mathbb{A}}
	\left (\frac{{\textstyle \sum_{k=1}^{n}}\left (  {\textstyle \prod_{j=1}^{N}} \alpha_{i_{j}}^{j}(k)  \right ) }{n}   \cdot
	\log\left ( \frac{\frac{{\textstyle \sum_{k=1}^{n}}\left (  {\textstyle \prod_{j=1}^{N}} \alpha_{i_{j}}^{j}(k)  \right ) }{n} }
	{ \textstyle \prod_{j=1}^{N} \frac{ {\textstyle \sum_{k=1}^{n} \alpha_{i_{j}}^{j}(k)} }{n}}    \right )    \right ) 
\end{gather}

\subsection{Limitations}\label{app:limitations}

\paragraph{Indeterminate Probability Theory.} As we summarized in Section~\ref{sec:summary}, we do not find any exceptions for our proposed three
conditional mutual independency axioms, see Axiom~\ref{axm:A_X_independence} Axiom~\ref{axm:Y_A_independence} and Axiom~\ref{axm:Y_X_independence}. 
And our proposed equation is derived from these axioms, in our opinion, this equation can be applied to any general random experiment.

\paragraph{IPNN.} IPNN is one neural network framework based on indeterminate probability theory, it has three limitations: 
(1) The split shape need to be predefined, a proper sample space for an unknown dataset can only be found with try and error. 
The latent variables are continuous in CIPNN~\cite{cipnn}, therefore this issue does not exist in CIPNN.
(2) It sometimes converges to local minimum, but we can avoid this problem with a proper model weights initialization, as discussed in Appendix~\ref{app:local_convergence}.
(3) As joint sample space increases exponentially, the memory consumption and computation time also increase accordingly. This issue only exist during training, 
and can be avoided through monte carlo method for prediction task, as discussed in CIPNN~\cite{cipnn}, this paper will not further discuss it.

\subsection{Pseudo Code Pytorch implementation of IPNN}\label{app:pseudocode}

See below:

\begin{lstlisting}
'''
Pseudo code of calculation of the loss and the inference posterior P^{A}(Y|X).

b                    --> batch size
y                    --> number of classification classes
[M_1, M_2, ..., M_N] --> split shape

inputs:
   logits: [b, M_1 + M_2 +, ..., M_N] # neural network outputs
   y_true: [b,y] # labels
outputs: 
   probability: [b,y] # the inference posterior P^{A}(Y|X)
   loss
'''

EINSUM_CHAR = 'ijklmnopqrstuvwIJKLMNOPQRSTUVW' # no special meaning.

logits = torch.split(logits, split_shape, dim = -1) 
# Shape of variables: [[b, M_1], [b, M_2], ..., [b, M_N]]
variables = [torch.softmax(_,dim = -1) for _ in logits]

# Joint sample space calculation
# Shape of joint_variables: [b, M_1, M_2, ..., M_N]
for i in range(len(variables)): 
if i == 0 : 
      joint_variables = variables[i]
else: 
      r_ = EINSUM_CHAR[:joint_variables.dim()-1]
      joint_variables = torch.einsum('b{},ba->b{}a'.format(r_,r_),joint_variables,variables[i]) # see (*@Equation~\ref{eq:P_A_X}@*) 


# OBSERVATION PHASE
r_ = EINSUM_CHAR[:joint_variables.dim()-1]
num_y_joint_current = torch.einsum('b{},by->y{}'.format(r_,r_),joint_variables,y_true) #  # see (*@Equation~\ref{eq:hh}@*) 
num_joint_current = torch.sum(joint_variables,dim = 0) # see (*@Equation~\ref{eq:gg}@*) 

# numerator and denominator of conditional probability P(Y|A^1,A^2,...,A^N)
num_y_joint += num_y_joint_current # see (*@Equation~\ref{eq:H}@*) 
num_joint += num_joint_current # see (*@Equation~\ref{eq:G}@*) 

# Shape of prob_y_joint: [y, M_1, M_2, ..., M_N]
prob_y_joint = num_y_joint / num_joint # see (*@Equation~\ref{eq:P_Y_A_2}@*) 


# INFERENCE PHASE
# Shape of probability: [b,y]
r_ = EINSUM_CHAR[:joint_variables.dim()-1]
probability = torch.einsum('y{},b{}->by'.format(r_,r_),prob_y_joint,joint_variables) # see (*@Equation~\ref{eq:P_Y_X_via_A_Train}@*) 


# loss function
loss = cross_entropy_loss(probability,y_true) # see (*@Equation~\ref{eq:loss}@*) 

\end{lstlisting}

\end{appendix}

\end{document}